\pdfoutput=1

\documentclass[11pt]{article}

\usepackage[final]{acl}

\usepackage{amsmath}
\usepackage{amssymb}
\usepackage{times}
\usepackage{latexsym}
\usepackage{kotex}
\usepackage{booktabs}
\usepackage{multirow}
\usepackage{lscape}
\usepackage{booktabs}
\usepackage{array}
\usepackage{longtable}
\usepackage{color}
\usepackage{tabularray}
\usepackage{arydshln}
\usepackage{subcaption}

\usepackage{tikz}
\usetikzlibrary{arrows.meta}  

\usepackage[T1]{fontenc}

\usepackage[utf8]{inputenc}

\usepackage{microtype}

\usepackage{inconsolata}

\usepackage{graphicx}
\usepackage{arydshln}

\title{CLEAR: Cross-Lingual Enhancement in Retrieval via Reverse-training}
\author{
    Seungyoon Lee$^{1}$, Minhyuk Kim$^{1}$, Seongtae Hong$^{1}$, Youngjoon Jang$^{1}$, \\ 
    \textbf{Dongsuk Oh$^{2}$, Heuiseok Lim$^{1}$\thanks{Corresponding author}} \\
    $^{1}$Department of Computer Science and Engineering, Korea University \\
    $^{2}$Department of English Language and Literature, Kyungpook National University \\
    \texttt{\{dltmddbs100, mhkim0929, ghdchlwls123, dew1701, limhseok\}@korea.ac.kr}, \\
    \texttt{inow3555@knu.ac.kr} \\
}

\begin{document}
\maketitle
\begin{abstract}
Existing multilingual embedding models often encounter challenges in cross-lingual scenarios due to imbalanced linguistic resources and less consideration of cross-lingual alignment during training. Although standardized contrastive learning approaches for cross-lingual adaptation are widely adopted, they may struggle to capture fundamental alignment between languages and degrade performance in well-aligned languages such as English. To address these challenges, we propose \textbf{C}ross-\textbf{L}ingual \textbf{E}nhancement in Retriev\textbf{A}l via \textbf{R}everse-training~(\textbf{CLEAR}), a novel loss function utilizing a reverse training scheme to improve retrieval performance across diverse cross-lingual retrieval scenarios. CLEAR leverages an English passage as a bridge to strengthen alignments between the target language and English, ensuring robust performance in the cross-lingual retrieval task. Our extensive experiments demonstrate that CLEAR achieves notable improvements in cross-lingual scenarios, with gains up to 15\%, particularly in low-resource languages, while minimizing performance degradation in English. Furthermore, our findings highlight that CLEAR offers promising effectiveness even in multilingual training, suggesting its potential for broad application and scalability. We release the code at \url{https://github.com/dltmddbs100/CLEAR}.
\end{abstract}

\section{Introduction}
The recent progress in Large Language Models~(LLMs) has enabled multilingual applications such as question answering via retrieval-augmented generation~(RAG), substantially increasing the demand for robust cross-lingual information retrieval system~\cite{lewis2020retrieval, Siriwardhana2022ImprovingTD,li-etal-2022-learning-cross, gao2023retrieval, zhang-etal-2023-miracl, kamalloo-etal-2023-evaluating, chirkova-etal-2024-retrieval, wang2024improving, wang2025speculative}. 

Nevertheless, existing multilingual embedding models widely used for information retrieval often suffer from imbalanced linguistic distribution and insufficient attention to cross-lingual alignment during training~\cite{izacard2021unsupervised,chen-etal-2024-m3,wang2024multilingual, zhang-etal-2024-mgte, sturua2024jina}. This leads to biased representations and suboptimal retrieval performance, particularly in low-resource languages~\cite{palta2022investigating,huang2023improving,yang-etal-2024-language-bias,hong2026improving}, exacerbating disparity in cross-lingual information retrieval scenarios.

\begin{figure}[t]
\centering 
\includegraphics[width=\linewidth]{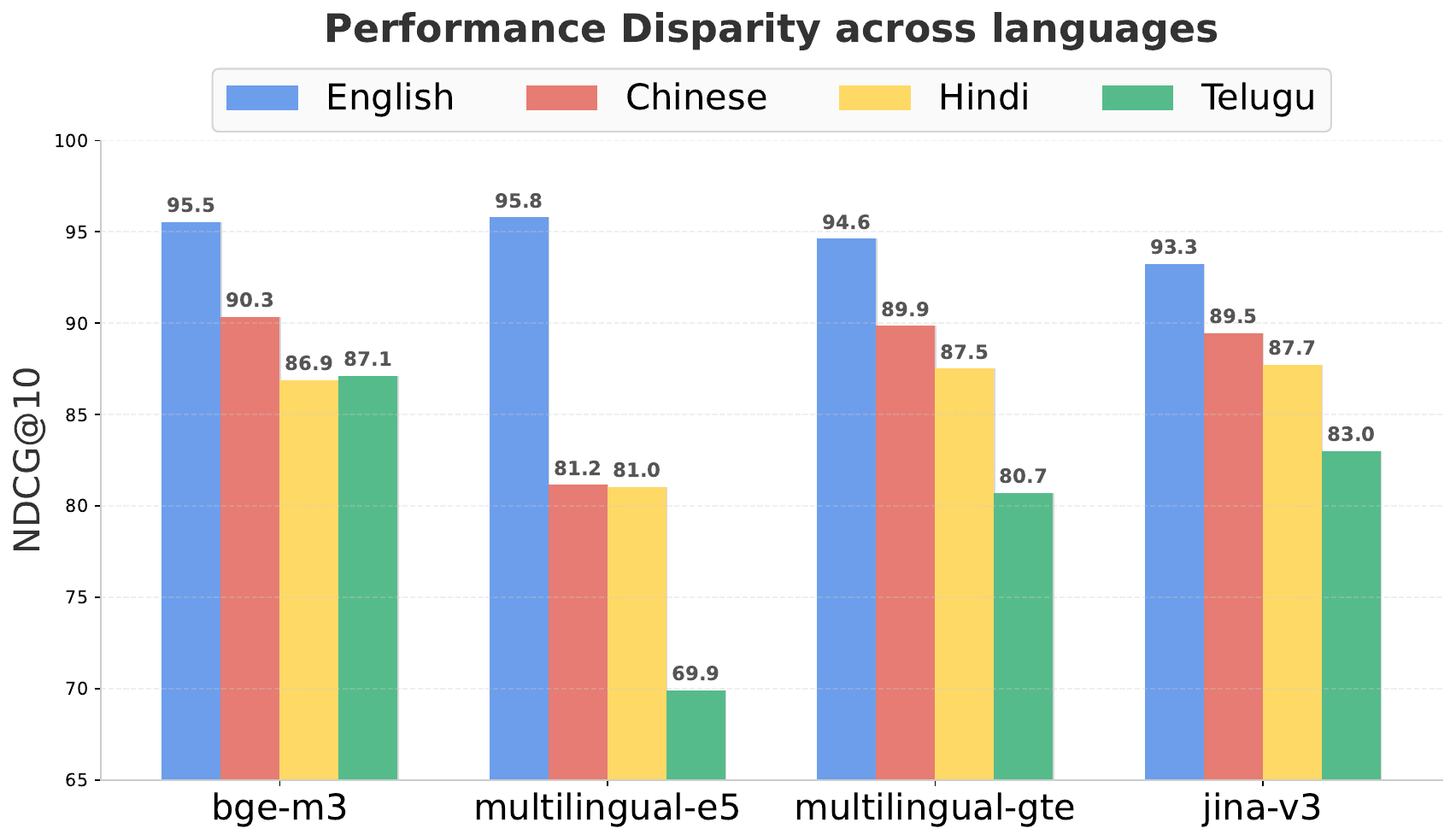}
\caption{Performance disparity of various embedding models across languages in a cross-lingual setup where English passage with other language queries in Belebele benchmark.}
\label{fig:disparity} 
\end{figure}

This disparity leads to lower performance due to diminished expressiveness in particular languages. As illustrated in Figure~\ref{fig:disparity}, user queries in low-resource languages lead to a considerable drop in retrieval performance due to the constraints of the model's representative capability. This increases the risk of providing users with inaccurate information, thereby posing a considerable challenge to ensuring equitable access to multilingual information~\cite{lawrie2023neural,park2025investigating}. This limitation not only hinders the accuracy and fairness of cross-lingual retrieval systems but also highlights the need for training strategies that explicitly enhance cross-lingual alignment, especially in resource-scarce contexts.

One of the prevalent approaches to address this issue involves enhancing cross-lingual alignment through an additional contrastive learning stage~\cite{shuaibo-etal-2022-supervised,wang-etal-2022-english,zhang-etal-2024-mgte,chen-etal-2024-m3}. The commonly adopted InfoNCE~\cite{oord2018representation} loss minimizes the distance between queries and gold passages while increasing the distance to negative samples. Although this loss function can be effective in encouraging query-passage similarity, it primarily focuses on distinguishing relevant passages based solely on the query. Consequently, it often captures only superficial representation ability and may fail to ensure fundamental alignment between different languages. Furthermore, these training strategies can degrade monolingual performance in the dominant language~(e.g., English) during cross-lingual training.

In this paper, we propose Cross-Lingual Enhancement in Retrieval via Reverse-training~(CLEAR), a novel training objective designed to improve retrieval performance across all cross-lingual scenarios where queries and passages are in different languages. CLEAR jointly trains on English and cross-lingual alignment by leveraging the English passage as a bridge to the target language, promoting diverse interactions among various components. In contrast to conventional methods that train models to retrieve related passages in response to a query, we introduce a reverse training scheme that fosters the model to capture multifaceted representation capabilities. This approach strengthens the foundational alignment between the languages, ensuring robustness across all cross-lingual scenarios.

Across extensive experiments spanning nine languages, our empirical findings demonstrate that CLEAR achieves substantial improvements in cross-lingual scenarios, especially exhibiting notable performance in low-resource languages, while concurrently mitigating the degradation of original proficiency in English. Furthermore, our approach substantiates its validity in the multilingual configuration where multiple languages are jointly trained. Our contributions are as follows:

\begin{itemize}
    \item We propose a novel cross-lingual specialized loss, CLEAR, that leverages a reverse training scheme based on English passage bridge to enhance cross-lingual capability.

    \item We empirically verify the effectiveness of CLEAR for cross-lingual retrieval tasks through extensive experiments using various embedding models on a range of high- and low-resource languages while minimizing the degradation of English performance compared to the standard training approach.
    
    \item We show that CLEAR extends beyond cross-lingual scenarios, also proving highly effective in multilingual training when multiple target languages are concurrently addressed.

\end{itemize}

\section{Related Work}
In the field of cross-lingual retrieval, existing studies can largely be categorized into two primary directions. The first direction employs translated pairs for direct fine-tuning to adapt retrieval models to target languages~\cite{litschko2018unsupervised,shi-etal-2021-cross,shuaibo-etal-2022-supervised,zhang-misra-2022-machine,zhuang2023augmenting}. For example, \citet{shi-etal-2021-cross} and \citet{zhuang2023augmenting} utilize the query generation model based on the translation of English query-passage pairs to generate synthetic queries, followed by the training of retriever on this dataset. 

On the other hand, the other line of research centers on knowledge distillation methods focusing on distilling insights from monolingual models into multilingual frameworks~\cite{reimers2020makingmonolingualsentenceembeddings,limkonchotiwat2022cl,li2022learning,Huang_2023,yang2024translatedistilllearningcrosslanguagedense,zhang2024jasper}. \citet{Huang_2023} introduces the Optimal Transport Distillation strategy to facilitate the transfer of knowledge from high to low resource languages by utilizing a well-trained monolingual retrieval model. Similarly, other studies distill representation or ranking knowledge of well-aligned language models into student models using parallel query-document pairs~\cite{li2022learning,limkonchotiwat2022cl,yang2024translatedistilllearningcrosslanguagedense}.

More recently, the focus has shifted toward training retrieval models on large-scale multilingual query-document datasets to map language-specific representations into a shared embedding space~\cite{chen-etal-2024-m3,zhang-etal-2024-mgte,sturua2024jina,wang2024multilingual,lee2025gemini}. 

While these approaches have succeeded in scaling coverage to a broader set of languages, they typically depend on vast amounts of parallel data and are often limited to capturing shallow cross-lingual interactions due to their reliance on the conventional InfoNCE loss. In this paper, we introduce a cross-lingual specialized loss function based on shared English passages to establish connections among languages. Our approach promotes diverse interactions among the components to enhance cross-lingual alignment in a resource-constrained environment.

\section{CLEAR}
We design CLEAR as a cross-lingual training loss function that leverages a reversal scheme to induce robust alignment between English and the target language.

\subsection{Overview of InfoNCE}
In general, retrievers learn meaningful representations based on similarity to identify the gold passage relevant to a given user query from a large passage pool. The prevalent approaches employ the InfoNCE loss~\cite{oord2018representation} with multiple negatives. The formula for the retrieval task is as follows.

Given a text pairs $(q_i, p_i^+)$, we assign a negative passage $p_{ij}^{-}$ for the i-th example:
\begin{align}
\mathcal{L}_{NCE} = - \frac{e^{\text{sim}(q_i,p_i^+)/\tau}}{e^{\text{sim}(q_i,p_i^+)/\tau} + \sum_{j} e^{\text{sim}(q_i,p_{ij}^-)/\tau}}
\label{eq:infonce}
\end{align}
This formula promotes the model to distinguish between related pairs~($q_i$, $p_i^+$) and unrelated passages~($p_{ij}^-$) within the embedding space, as quantified by a cosine similarity $sim$. 
Based on the InfoNCE loss, we modify the loss to promote cross-lingual alignment to suit our goal.

\subsection{Proposed Strategy}
\begin{figure}[t]
\centering 
\includegraphics[width=0.9\linewidth]{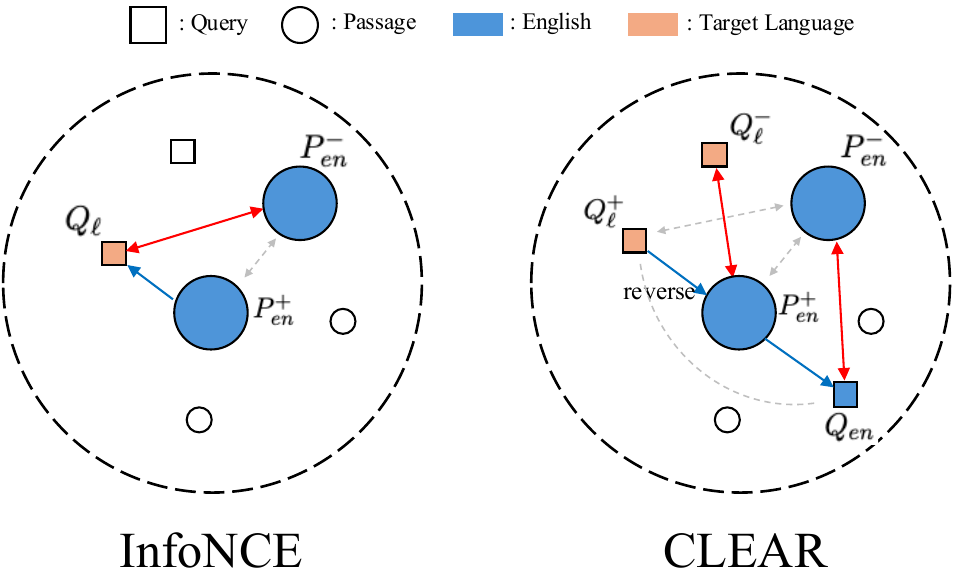}
\caption{Comparison of the core idea of CLEAR with the standard InfoNCE loss. Solid arrows indicate the direct effects of training, while dashed arrows represent the indirect interactions, such as the resulting attraction or repulsion between representations.}
\label{fig:clear} 
\end{figure}

CLEAR combines traditional contrastive learning with cross-lingual considerations to provide a solid foundation for enhancing retrieval capabilities. In Figure~\ref{fig:clear}, a comparison of the interactions between queries and passages induced by conventional InfoNCE and CLEAR during training is shown. Compared to InfoNCE, CLEAR achieves sophisticated cross-lingual alignment by establishing diverse direct and indirect interactions centered around \(P^{+}_{en}\).

CLEAR consists of a universal InfoNCE term designed for learning English representations alongside a cross-lingual term that encourages alignment between the target language ($\ell$) and English ($en$). We define our overall loss function as follows:
\begin{equation}
\begin{aligned}
\mathcal{{L}_{\text{CLEAR}}} = \lambda_1 \cdot \mathcal{L}_{\text{NCE}_{en}} + \lambda_2 \cdot \mathcal{L}_{\text{CL}} + \lambda_3 \cdot \mathcal{L}_{\text{KL}}
\end{aligned}
\end{equation}

\(\mathcal{L}_{\text{NCE}_{en}}\) represents the standard NCE loss associated with English pairs~(Equation~\ref{eq:infonce}), aiming at preserving the model's inherent performance in English while providing a language bridge via the passage. The direct training signal of cross-lingual alignment occurs through \(\mathcal{L}_{\text{CL}}\), which is the reversed cross-lingual loss term to align English with the target language:
\begin{equation}
\begin{aligned}
& \mathcal{L}_{\text{CL}}(p_{{en_{i}}}, q_{\ell_{i}}^+, q_{\ell_{ij}}^-) = -\frac{e^{z_i^+/{\tau}}}{e^{z_i^+/{\tau}} + \sum_{j} e^{z_{ij}^-/{\tau}}} 
\end{aligned} 
\label{eq:3} 
\end{equation}
where \( z_i^+ \) and \( z_{ij}^- \) is defined as:
\begin{equation}
\begin{aligned}
z_i^+ = \text{sim}(p_{{en_{i}}}, q_{\ell_{i}}^+), \; z_{ij}^- = \text{sim}(p_{{en_{i}}}, q_{\ell_{ij}}^-)
\end{aligned}
\end{equation}
where $p_{en_{i}}$ denotes the English passage that serves as an anchor, while $q_{\ell_{i}}^+$ and $q_{\ell_{ij}}^-$ refer to the positive and negative target language queries corresponding to $p_{en_{i}}$. Thus, the training objective induces the model to pull the gold query closer to $p_{en_{i}}$ and push unrelated queries further.

Next, \(\mathcal{L}_{\text{KL}}\) denotes the KL-Divergence~\cite{kullback1951information} between the similarity matrices $S_{en}$ and $S_{CL}$:
\begin{equation} 
\text{KL}(S_{en} \parallel S_{CL}) = \sum_{i,j} S_{en}[i, j] \log \frac{S_{en}[i, j]}{S_{CL}[i, j]}
\end{equation}
where \(S_{en}[i, j] = \text{sim}(q_{en_i}, p_{en_j})\) and 
\(S_{CL}[i, j] = \text{sim}(p_{en_j}, q_{\ell_i})\) in the batch.
In this manner, we harmonize similarity distributions between language pairs to support consistent representation spaces. The detailed strategies of CLEAR are as follows.

\paragraph{Reversal Scheme} We introduce a reversal scheme in \(\mathcal{L}_{\text{CL}}\) that swaps the roles of the query and passage to provide a new perspective on cross-lingual training signals. Unlike standard approaches, which consider $(q_{{en_{i}}}, p_{\ell_{i}}^+, p_{\ell_{ij}}^-)$ as anchor, positive and negative respectively, this scheme encourages the model to use passage as an anchor and maximize the similarity with the corresponding gold target language query as shown in Equation~\ref{eq:3}.

This scheme enables the utilization of unrelated target language queries from in-batch samples as negative samples during training, thereby facilitating contrastive learning. By offering a new direction of training signal beyond the standard training signal, the reversal scheme supports the model to learn from passage-to-query perspectives, thereby promoting more robust cross-lingual adaptation.

\paragraph{Passage Bridge} We share the same English passage~$p_{en_{i}}$ for both the target language query and corresponding English query. This approach creates a bridge between languages by jointly optimizing relevant query-passage pairs and achieving distributional alignment. As illustrated by the dotted arrows in Figure~\ref{fig:clear}, $Q_{\ell}^{+}$ moves closer to $Q_{en}$, which conveys the same meaning, and further from the negative English passage, due to the interaction rise via the passage bridge. This strategy promotes the model to consider a broader range of mutual interactions during training, facilitating robust alignment between language representations.

\paragraph{Distribution Approximation} Relying solely on instance-level contrastive signals may lead to less robust or more fragmented representations. To mitigate this issue, we employ a KL-Divergence loss to align the similarity distribution between $P_{en}$ and $Q_{\ell}$ with that between $P_{en}$ and $Q_{en}$. While other loss terms operate at the instance level by optimizing query-passage pairs, the KL term goes beyond individual point-to-point relationships; it shapes the overall semantic topology, ensuring that the global organization of meanings remains coherent across languages. This mechanism complements mutual interactions among the loss components and encourages the model to maintain shared representations across English and the target language.

\paragraph{} Through the composite design, CLEAR benefits from a strong alignment signal between the target language and English while preserving the model's proficiency in English during the training.

\section{Experimental Setup}
\subsection{Training}
\paragraph{Dataset}
To enable cross-lingual training, target language queries paired with corresponding English passages are necessary. We employ the English portion of the MIRACL~\cite{zhang-etal-2023-miracl} training set and MLQA~\cite{lewis2020mlqa}, both of which provide queries mapped to gold passages. To gain parallel queries with English, we use the NLLB~\cite{costa2022no} translation model\footnote{\url{https://huggingface.co/facebook/nllb-200-3.3B}}. We exclude queries sharing the same passage to prevent false negatives that may arise from duplicate passages within a single batch, resulting in a collection of unique query-passage pairs. This process yields 12,698 training samples for each language.

Regarding hard-negative selection, following the findings of \citet{gabriel2024nv}, which reports that selecting top-k candidates within the range of 30 to 100 effectively reduces false negatives during the negative mining, we sample 5 hard negatives for each training sample using each embedding model. Further details and exact training example can be found in Appendix~\ref{appen:hyper} and \ref{appen:train_ex}.
\begin{table*}[htb!]
\centering
\scalebox{0.65}{
\begin{tabular}{cccccccccc}
\toprule
\multirow{2}{*}{\textbf{Model}} & \multirow{2}{*}{\textbf{Language}} & \multicolumn{2}{c}{\textbf{English-English}} & \multicolumn{3}{c}{\textbf{English-Lang}} & \multicolumn{3}{c}{\textbf{Lang-English}} \\
\cmidrule(lr){3-4}
\cmidrule(lr){5-7}
\cmidrule(lr){8-10}
& & \textbf{InfoNCE} & \textbf{CLEAR} & \textbf{Base} & \textbf{InfoNCE} & \textbf{CLEAR} & \textbf{Base} & \textbf{InfoNCE} & \textbf{CLEAR} \\

\midrule[1.2pt]
\multicolumn{10}{c}{\textit{\textbf{Belebele}}}\\
\midrule[1.2pt]

\multirow{7}{*}{bge-m3}
& zh & 94.65 (-0.90) & \textbf{95.47 (-0.08)} & 90.35 & 91.69 & \textbf{92.64} & 89.81 & 91.63 & \textbf{92.02} \\
& es & 95.51 (-0.04) & \textbf{96.14 (+0.59)} & 91.99 & 92.81 & \textbf{93.39} & 92.05 & 93.50 & \textbf{93.82} \\
& de & 95.61 (+0.06) & \textbf{96.11 (+0.56)} & 92.81 & 93.98 & \textbf{94.58} & 92.90 & 94.30 & \textbf{94.52} \\
& hi & 95.15 (-0.40) & \textbf{95.75 (+0.20)} & 86.90 & 89.14 & \textbf{90.16} & 90.39 & 92.86 & \textbf{93.16} \\
& vi & 95.14 (-0.41) & \textbf{95.79 (+0.24)} & 92.58 & 93.04 & \textbf{93.12} & 93.04 & 92.58 & \textbf{93.54} \\
& te & 94.93 (-0.62) & \textbf{95.74 (+0.19)} & 87.12 & 89.07 & \textbf{90.12} & 89.19 & 92.08 & \textbf{92.74} \\
& bn & 95.18 (-0.37) & \textbf{95.81 (+0.26)} & 87.58 & 89.79 & \textbf{90.85} & 90.17 & 92.37 & \textbf{93.08} \\

\midrule

\multirow{7}{*}{multilingual-e5}
& zh & 94.18 (-1.61) & \textbf{95.06 (-0.73)} & 81.16 & 87.00 & \textbf{88.89} & 85.49 & 90.02 & \textbf{90.75} \\
& es & 94.38 (-1.41) & \textbf{95.40 (-0.39)} & 89.46 & 89.84 & \textbf{91.61} & 92.43 & 92.15 & \textbf{93.13} \\
& de & 94.49 (-1.30) & \textbf{95.06 (-0.73)} & 90.77 & 90.97 & \textbf{91.94} & 92.16 & 92.68 & \textbf{92.86} \\
& hi & 94.11 (-1.68) & \textbf{95.28 (-0.51)} & 81.05 & 83.11 & \textbf{86.10} & 88.44 & 91.08 & \textbf{92.13} \\
& vi & 94.22 (-1.57) & \textbf{95.07 (-0.72)} & 86.03 & 87.11 & \textbf{88.63} & 88.58 & 91.81 & \textbf{91.95} \\
& te & 93.80 (-1.99) & \textbf{95.11 (-0.68)} & 69.90 & 77.14 & \textbf{80.97} & 84.88 & 88.05 & \textbf{88.99} \\
& bn & 93.83 (-1.96) & \textbf{95.28 (-0.51)} & 75.55 & 81.81 & \textbf{85.44} & 84.31 & 88.91 & \textbf{89.92} \\

\midrule

\multirow{7}{*}{gte-multilingual}
& zh & 94.38 (-0.25) & \textbf{95.15 (+0.52)} & 89.86 & 92.30 & \textbf{92.67} & 91.51 & 92.05 & \textbf{92.80}\\
& es & 95.51 (+0.88) & \textbf{95.73 (+1.10)} & 91.71 & 92.86 & \textbf{93.28} & 90.20 & 93.55 & \textbf{93.87} \\
& de & 95.32 (+0.69) & \textbf{95.67 (+1.04)} & 91.21 & 92.92 & \textbf{93.26} & 89.42 & 93.45 & \textbf{93.95} \\
& hi & 94.93 (+0.30) & \textbf{95.40 (+0.77)} & 87.51 & 89.34 & \textbf{89.96} & 89.55 & 92.45 & \textbf{93.05} \\
& vi & 95.02 (+0.39) & \textbf{95.71 (+1.08)} & 89.37 & 91.59 & \textbf{92.23} & 90.48 & 93.14 & \textbf{93.52} \\
& te & 94.63 (0.00) & \textbf{95.32 (+0.69)} & 80.70 & 84.77 & \textbf{86.31} & 88.46 & 89.96 & \textbf{90.92} \\
& bn & 94.54 (-0.09) & \textbf{95.40 (+0.77)} & 82.13 & 85.18 & \textbf{86.64} & 87.81 & 91.03 & \textbf{92.14} \\

\midrule

\multirow{7}{*}{jina-v3}
& zh & 94.97 (+1.71) & \textbf{95.31 (+2.05)} & 89.46 & 91.90 & \textbf{92.67} & 89.64 & 91.74 & \textbf{92.18} \\
& es & 95.21 (+1.95) & \textbf{95.66 (+2.40)} & 91.40 & 93.57 & \textbf{94.29} & 92.64 & \textbf{93.88} & 93.86 \\
& de & 95.29 (+2.03) & \textbf{95.62 (+2.36)} & 91.75 & 93.88 & \textbf{94.46} & 92.39 & 94.12 & \textbf{94.46} \\
& hi & 94.01 (+0.75) & \textbf{95.59 (+2.33)} & 87.74 & 89.90 & \textbf{90.70} & 91.50 & 93.02 & \textbf{93.34} \\
& vi & 94.01 (+0.75) & \textbf{95.50 (+2.24)} & 90.26 & \textbf{92.68} & 92.63 & 90.98 & 93.12 & \textbf{93.40} \\
& te & 91.33 (-1.93) & \textbf{95.69 (+2.43)} & 83.02 & 85.57 & \textbf{87.32} & 88.99 & 91.87 & \textbf{92.58} \\
& bn & 94.01 (+0.75) & \textbf{95.75 (+2.49)} & 86.56 & 89.22 & \textbf{90.89} & 91.14 & \textbf{93.52} & 93.43 \\

\midrule

Total & Average & 94.58 (-0.22) & \textbf{95.52 (+0.71)} & 87.00 & 89.36 & \textbf{90.56} & 89.95 & 92.18 & \textbf{92.72} \\

\midrule[1.2pt]
\multicolumn{10}{c}{\textit{\textbf{XQuAD}}}\\
\midrule[1.2pt]

\multirow{4}{*}{bge-m3}
& ar & 96.29 (-0.88) & \textbf{96.72 (-0.45)} & 92.24 & 93.08 & \textbf{93.50} & 92.38 & 94.41 & \textbf{94.71} \\
& zh & 96.13 (-1.04) & \textbf{96.70 (-0.47)} & 94.04 & 94.03 & \textbf{94.68} & 93.18 & 94.21 & \textbf{94.82} \\
& es & 96.80 (-0.37) & \textbf{96.91 (-0.26)} & 96.14 & 95.97 & \textbf{96.14} & 95.96 & \textbf{96.42} & 96.34 \\
& ru & 96.24 (-0.93) & \textbf{96.91 (-0.26)} & 95.58 & 95.13 & \textbf{95.83} & 94.57 & 94.98 & \textbf{95.16} \\

\midrule

\multirow{4}{*}{multilingual-e5}
& ar & 94.70 (-3.32) & \textbf{96.23 (-1.79)} & 87.29 & 87.58 & \textbf{89.83} & 91.22 & 91.69 & \textbf{92.61} \\
& zh & 95.22 (-2.80) & \textbf{96.01 (-2.01)} & 89.60 & 90.64 & \textbf{91.71} & 91.02 & 92.44 & \textbf{93.44} \\
& es & 95.51 (-2.51) & \textbf{95.94 (-2.08)} & \textbf{96.15} & 93.39 & 94.00 & \textbf{96.41} & 94.42 & 94.62 \\
& ru & 95.18 (-2.84) & \textbf{95.94 (-2.08)} & \textbf{93.06} & 91.90 & 92.54 & \textbf{93.22} & 92.54 & 93.11 \\

\midrule

\multirow{4}{*}{gte-multilingual}
& ar & 94.70 (-3.32) & \textbf{96.23 (-1.79)} & 87.29 & 87.58 & \textbf{89.83} & 91.22 & 91.69 & \textbf{92.61} \\
& zh & 96.84 (-1.09) & \textbf{97.43 (-0.50)} & 93.98 & 94.27 & \textbf{94.87} & 92.07 & 93.67 & \textbf{94.05} \\
& es & 97.56 (-0.37) & \textbf{97.59 (-0.34)} & 96.14 & 95.88 & \textbf{96.45} & 95.79 & 96.53 & \textbf{96.66} \\
& ru & 97.04 (-0.89) & \textbf{97.40 (-0.53)} & 94.38 & 94.06 & \textbf{94.57} & 94.24 & 94.88 & \textbf{95.21} \\

\midrule

\multirow{4}{*}{jina-v3}
& ar & \textbf{97.24 (+1.15)} & 97.16 (+1.07) & 90.58 & 93.25 & \textbf{93.90} & 93.21 & 95.02 & \textbf{95.19} \\
& zh & 97.31 (+1.22) & \textbf{97.33 (+1.24)} & 92.65 & 94.72 & \textbf{95.05} & 92.79 & 95.36 & \textbf{95.75} \\
& es & \textbf{97.36 (+1.27)} & 97.17 (+1.08) & 94.76 & \textbf{96.00} & 95.92 & 96.12 & 96.99 & \textbf{97.02} \\
& ru & \textbf{97.25 (+1.16)} & 97.19 (+1.10) & 93.97 & 95.88 & \textbf{96.06} & 94.38 & 95.53 & \textbf{95.82} \\

\midrule

Total & Average & 96.47 (-0.84) & \textbf{96.88 (-0.44)} & 93.00 & 93.40 & \textbf{94.05} & 93.65 & 94.52 & \textbf{94.89} \\

\bottomrule
\end{tabular}}
\caption{Comprehensive cross-lingual evaluation results. In each cross-lingual setting, `Lang' refers to the target language. The first word denotes the language of the passage, and the second one denotes the language of the query. The value in `()' indicates the difference in performance from the original model.}
\label{tab:main_bele}
\end{table*}


\paragraph{Models}
We adopt four widely used multilingual embedding models across various tasks: bge-m3~\cite{chen-etal-2024-m3}, multilingual-e5~\cite{wang2024multilingual}, gte-multilingual~\cite{zhang-etal-2024-mgte}, and jina-v3~\cite{sturua2024jina}. The models are trained under the identical hyper-parameters and evaluations are conducted under the same conditions. We heuristically set weights for each loss term in our experiments: $\lambda_1 = 0.4$, $\lambda_2 = 0.4$, and $\lambda_3 = 0.2$. Regarding this, the sensitivity analysis of loss weight parameters is provided in Appendix~\ref{appen:loss_param}.

\subsection{Baseline}
We employ the standard InfoNCE loss, which integrates in-batch negative sampling with external multiple negatives~\cite{henderson2017efficient} as a major baseline for our experiments. To be specific, we train the model to increase the similarity between a target language query~(used as the anchor) and its relevant English passage, focusing exclusively on the cross-lingual alignment.

To ensure fair comparison, we utilize the same five hard negative samples for target language queries. We also conduct query negative mining for our cross-lingual term, which leverages queries as negative samples by calculating similarity scores between the gold passage and queries.

\subsection{Evaluation}
\paragraph{Language Scope}
We perform downstream task evaluation on nine languages: Arabic~(ar), German~(de), Chinese~(zh), Russian~(ru), Spanish~(es), Hindi~(hi), Vietnamese~(vi), Telugu~(te) and Bengali~(bn). They were chosen to provide a mix of high-, medium-
and low-resource languages, typological and script diversity while satisfying the practical constraints of available evaluation datasets. We refer to German, Chinese, Russian and Spanish as high-resource, Arabic, Hindi and Vietnamese as medium-resource, and Telugu and Bengali as low-resource languages.

\paragraph{Cross-lingual Scenario}
We conduct a comprehensive evaluation across a wide range of cross-lingual scenarios. Since our aim covers cross-lingual evaluation across all directions ($P_{en}$ - $Q_{\ell}$ / $P_{\ell}$ - $Q_{en}$), the same question-passage pairs must exist in multiple languages to enable the evaluation of retrieval performance across different languages, making essential to employ datasets that are fully parallel between English and target languages. To this end, we employ two cross-lingual retrieval benchmarks: Belebele~\cite{bandarkar-etal-2024-belebele}, which covers 122 language variants including English, and XQuAD~\cite{artetxe-etal-2020-cross}, which includes 11 languages. Both benchmarks are included in the authorized evaluation framework MMTEB tasks~\cite{enevoldsen2025mmteb}, which are driven by the expansion of MTEB~\cite{muennighoff2023mteb}. As a part of MTEB, these are widely adopted evaluation datasets in contemporary retrieval works~\cite{chen-etal-2024-m3,sturua2024jina,lee2025gemini,zhang2025qwen3}. More details about benchmarks can be found in Appendix~\ref{appen:benchmark}.

For each language present in both benchmarks, we evaluate on both Belebele and XQuAD; languages exclusive to Belebele are evaluated only with that benchmark. Furthermore, we assess the preservation of the model’s English retrieval capabilities by measuring the difference in English performance after the cross-lingual training. We use nDCG@10~\cite{jarvelin2002cumulated} as the primary evaluation metric.

\paragraph{Multilingual Expansion}
We also evaluate CLEAR in a multilingual setup where multiple languages are concurrently learned. We construct a multilingual training set by combining 1,410 non-overlapping samples per language from a cross-lingual training dataset. Then we train the model on this combined dataset. By assessing performance in both cross-lingual and target-language-only scenarios across all languages considered in our experiments, we demonstrate the scalability of CLEAR to multilingual training scenarios.

\section{Results}
We first investigate the effects of CLEAR on cross-lingual scenarios under various languages and embedding models in Section~\ref{sec:5.1}. Then in Section~\ref{sec:5.2}, we examine the generalization of CLEAR within a multilingual training setup involving nine mixed languages. Finally, by conducting an ablation study of CLEAR's core strategies in Section~\ref{sec:5.3}, we demonstrate the validity of our approach. A further experiment report on all languages and metrics is available in Appendix~\ref{appen:all_results}.

\subsection{Comprehensive Cross-lingual Retrieval}
\label{sec:5.1}
Table~\ref{tab:main_bele} shows the effectiveness of CLEAR for cross-lingual adaptation, regardless of language or model. CLEAR consistently outperforms InfoNCE in most cases, regardless of whether the passage or query is in the target language.

\paragraph{Low-resource Languages}
Notably, this performance gap is more pronounced for low-resource languages. As shown in Table~\ref{tab:main_bele}, on Bengali and Telugu, CLEAR achieves scores of 80.97 and 85.44 with multilingual-e5, which exceed the original model by more than 13\% and surpass InfoNCE~(77.14 and 81.81) by 4 points in the English-Lang setup. Similarly, for gte-multilingual, CLEAR reaches 86.31 and 86.64, showing a greater margin compared to other high-resource languages.

This is related to the performance of the original models concerning target languages. Both multilingual-e5 and gte-multilingual yield the lowest scores for low-resource languages among all models. This implies that an imbalance in training data causes a representational gap between English and the target languages. While InfoNCE seeks to address this gap solely with respect to the target language, CLEAR leverages English passages as a bridge, allowing the model to share interactions between English queries and underrepresented target language queries. As a result, CLEAR generalizes well in low-resource languages.

\begin{table*}[ht]
\centering
\renewcommand{\arraystretch}{1.2}
\resizebox{0.8\textwidth}{!}{
\begin{tabular}{cccccccccccc}
\toprule
\textbf{Model} & \textbf{Passage-Query} & \textbf{Method} & \textbf{ar} & \textbf{de} & \textbf{zh} & \textbf{ru} & \textbf{es} & \textbf{hi} & \textbf{vi} & \textbf{te} & \textbf{bn} \\ 

\toprule

\multirow{4}{*}{multilingual-e5} & \multirow{2}{*}{English-Lang} & InfoNCE & 82.58 & 89.99 & 86.84 & 89.16 & 89.46 & 83.10 & 86.69 & 76.48 & 81.85 \\ 
 &  & CLEAR & \textbf{85.12} & \textbf{91.87} & \textbf{88.85} & \textbf{91.68} & \textbf{91.98} & \textbf{85.94} & \textbf{88.85} & \textbf{80.22} & \textbf{84.51} \\ 
 \cline{2-12}
 & \multirow{2}{*}{Lang-English} & InfoNCE & 88.08 & 92.29 & 90.02 & 91.39 & 91.58 & 90.97 & 90.44 & 88.63 & 88.93 \\ 
 &  & CLEAR & \textbf{88.89} & \textbf{93.47} & \textbf{90.78} & \textbf{92.25} & \textbf{93.14} & \textbf{92.39} & \textbf{91.47} & \textbf{89.50} & \textbf{90.68} \\ 

\midrule
 
\multirow{4}{*}{gte-multilingual} & \multirow{2}{*}{English-Lang} & InfoNCE & 86.38 & 92.12 & 91.92 & 91.80 & 92.44 & 89.13 & 91.26 & 83.53 & 84.42 \\
 &  & CLEAR & \textbf{87.77} & \textbf{92.92} & \textbf{92.70} & \textbf{92.56} & \textbf{93.29} & \textbf{90.13} & \textbf{91.82} & \textbf{85.19} & \textbf{85.70} \\ 
 \cline{2-12}
 & \multirow{2}{*}{Lang-English} & InfoNCE & 90.75 & 93.20 & 92.11 & 92.71 & 93.10 & 92.53 & 92.89 & 89.91 & 91.08 \\ 
 &  & CLEAR & \textbf{91.44} & \textbf{93.55} & \textbf{92.52} & \textbf{93.28} & \textbf{93.53} & \textbf{92.78} & \textbf{93.23} & \textbf{91.03} & \textbf{92.02} \\ 

\bottomrule
 
\end{tabular}}
\caption{Cross-lingual evaluation results in multilingual training setup in Belebele benchmark.}
\label{tab:multi_cross}
\end{table*}

\begin{table}[t]
\centering
\resizebox{0.8\linewidth}{!}{
\begin{tabular}{ccccc}

\toprule

\multirow{2}{*}{Language} & \multicolumn{2}{c}{multilingual-e5} & \multicolumn{2}{c}{gte-multilingual} \\
\cmidrule(lr){2-3} \cmidrule(lr){4-5}
& InfoNCE & CLEAR & InfoNCE & CLEAR \\
\midrule
en & 94.94 & \textbf{95.61} & 94.66 & \textbf{95.46} \\
ar & 86.82 & \textbf{88.42} & 88.37 & \textbf{89.10} \\
de & 91.15 & \textbf{93.27} & 92.89 & \textbf{93.47} \\
zh & 91.28 & \textbf{93.05} & 92.85 & \textbf{93.65} \\
ru & 92.15 & \textbf{93.09} & 93.07 & \textbf{93.83} \\
es & 90.91 & \textbf{92.51} & 92.55 & \textbf{93.44} \\
hi & 84.24 & \textbf{86.07} & 88.48 & \textbf{89.34} \\
vi & 90.77 & \textbf{92.43} & 92.18 & \textbf{93.08} \\
te & 80.67 & \textbf{82.81} & 83.71 & \textbf{84.86} \\
bn & 85.28 & \textbf{87.56} & 86.01 & \textbf{87.63} \\

\bottomrule

\end{tabular}}
\caption{Monolingual performances in multilingual training setup. Each row presents the result where the passage and query are in the same language.}
\label{tab:multi_mono}
\end{table}

\paragraph{Generalization in Lang-English Setup}
We observe that CLEAR remains effective even when the passage is presented in the target language~(Lang-English), a setting not accounted for during training. This stems from fundamental alignment. CLEAR does not merely learn target language expressiveness dependent on the query; rather, it enhances the fundamental representational ability between the target language and English by considering reverse directions, generalizing well to unseen target language passages. This suggests that CLEAR can be a robust training approach in real-world environments where target language passage corpora are limited for training the retriever and demonstrates its effectiveness in diverse cross-lingual scenarios.

\paragraph{Preservation of English Ability}
In general, training methods that target cross-lingual alignment inevitably have a negative impact on monolingual performance. Table~\ref{tab:main_bele} also shows that cross-lingual training affects monolingual performance in English. However, CLEAR achieves superior performance in cross-lingual scenarios while reducing the leakage of the model's inherent capabilities in English or even achieving improvements. CLEAR yields a total average score of 96.88 in English results in XQuAD, reflecting a smaller decrease compared to InfoNCE’s 96.47. Notably, CLEAR even improves performance for most models on the Belebele benchmark. In contrast, InfoNCE shows a greater decrease~(multilingual-e5) or a marginal increase~(jina-v3).

\begin{figure}[t]
\centering
\subcaptionbox{InfoNCE}{\includegraphics[width=0.9\linewidth]{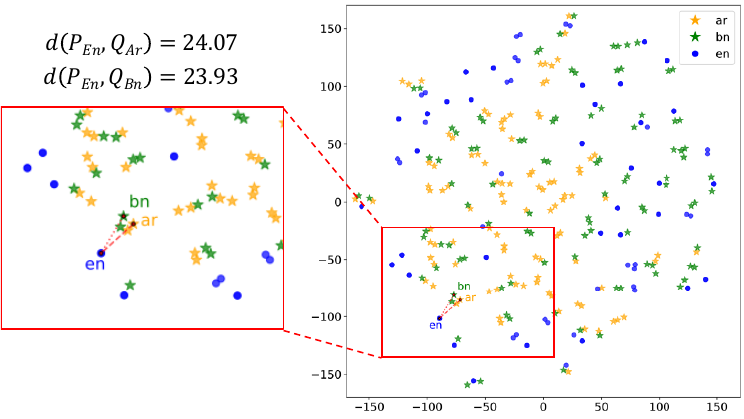}}
\vfill
\subcaptionbox{CLEAR}{\includegraphics[width=0.9\linewidth]{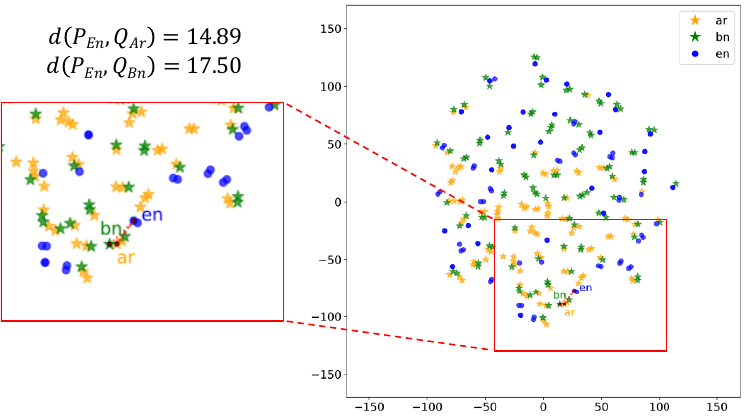}}
\vfill
\caption{T-SNE visualization of the embeddings for multilingual-e5 with English passage and Arabic, Bengali queries after multilingual training. We randomly select 100 pairs from the Belebele and measure the distance between the embeddings of identical gold pairs.}
\label{fig:multi_dist}
\end{figure}

This indicates that the strategy of CLEAR helps preserve the original English alignment while also considering the target language. \(\mathcal{L}_{\text{NCE}_\text{en}}\) encourages the model to maintain its existing English proficiency during the training, and the integration with \(\mathcal{L}_{\text{CL}}\) enables the model to consider the capability in the target language. Through this joint training approach, CLEAR offers a practical solution that can address real-world needs where both cross-lingual and English performance must be considered.

\begin{table*}[htb!]
\centering
\renewcommand{\arraystretch}{1.1}
\resizebox{\textwidth}{!}{
\begin{tabular}{lccccccccccc}
\toprule
\textbf{Method} & \textbf{ar} & \textbf{de} & \textbf{zh} & \textbf{ru} & \textbf{es} & \textbf{hi} & \textbf{vi} & \textbf{te} & \textbf{bn} & \textbf{Average} \\
\midrule



CLEAR$_{e5}$ & 83.78 / 88.23 & \textbf{91.94} / \textbf{92.86} & \textbf{88.89} / \textbf{90.75} & \textbf{90.59} / \textbf{92.16} & \textbf{91.61} / \textbf{93.13} & \textbf{86.10} / \textbf{92.13} & \textbf{88.63} / \textbf{91.95} & \textbf{80.97} / \textbf{88.99} & \textbf{85.44} / \textbf{89.92} & \textbf{87.55} / \textbf{91.12} \\

\hdashline[0.8pt / 1.2pt]
\; w/o \(\mathcal{L}_{\text{KL}}\) & \textbf{83.82} / \textbf{88.49} & 91.61 / 92.29 & 88.59 / 90.43 & 89.85 / 91.73 & 90.76 / 92.49 & 85.05 / 91.10 & 88.52 / 91.69 & 79.68 / 88.67 & 84.40 / 89.62 & 86.92 / 90.72 \\ 

\; w/o Reverse & 82.81 / 87.49 & 91.31 / 92.47 & 87.44 / 89.69 & 89.77 / 90.90 & 89.91 / 92.01 & 84.12 / 90.84 & 87.41 / 91.36 & 78.21 / 87.96 & 82.77 / 88.89 & 85.97 / 90.18 \\

\; w/o \(\mathcal{L}_{\text{NCE}_\text{en}}\) & 81.95 / 87.24 & 90.26 / 91.55 & 86.98 / 89.09 & 89.19 / 90.98 & 89.88 / 91.45 & 83.71 / 90.21 & 87.67 / 90.38 & 79.00 / 87.21 & 83.58 / 88.95 & 85.80 / 89.67 \\

\midrule

CLEAR$_{gte}$ & \textbf{87.91} / \textbf{91.46} & \textbf{93.26} / \textbf{93.95} & \textbf{92.67} / \textbf{92.80} & \textbf{92.94} / \textbf{93.33} & \textbf{93.28} / \textbf{93.87} & \textbf{89.96} / \textbf{93.05} & \textbf{92.23} / 93.52 & \textbf{86.31} / 90.92 & \textbf{86.64} / \textbf{92.14} & \textbf{90.58} / \textbf{92.78} \\

\hdashline[0.8pt/1.2pt]

\; w/o \(\mathcal{L}_{\text{KL}}\) & 87.13 / 91.16 & 92.90 / 93.70 & 92.60 / 92.48 & 92.50 / 93.28 & 92.92 / 93.46 & 89.59 / 92.89 & 91.80 / 93.27 & 85.01 / \textbf{91.18} & 85.79 / 91.70 & 90.03 / 92.57 \\

\; w/o Reverse & 87.19 / 90.93 & 93.01 / 93.74 & 92.46 / 92.35 & 92.54 / 93.32 & 93.16 / 93.63 & 89.73 / 92.68 & 91.67 / \textbf{93.62} & 85.08 / 91.01 & 85.91 / 91.47 & 90.08 / 92.53 \\

\; w/o \(\mathcal{L}_{\text{NCE}_\text{en}}\) & 84.36 / 88.29 & 90.96 / 91.30 & 89.59 / 89.93 & 90.17 / 90.65 & 91.46 / 92.10 & 86.37 / 90.67 & 89.79 / 90.87 & 81.41 / 87.42 & 85.17 / 90.77 & 87.70 / 90.22 \\

\bottomrule
\end{tabular}}
\caption{Ablation results (English-Lang / Lang-English) for key components in CLEAR on multilingual-e5 and gte.}
\label{tab:ablation}
\end{table*}

\subsection{Multilingual Training}
\label{sec:5.2}
We also find that CLEAR provides significant benefits in a multilingual configuration, where multiple languages are trained together. As shown in Table~\ref{tab:multi_cross}, CLEAR outperforms InfoNCE by a large margin across all languages.

Furthermore, despite targeting cross-lingual retrieval, CLEAR remains highly effective in a monolingual setup. Table~\ref{tab:multi_mono} shows monolingual retrieval results after the multilingual training, where CLEAR consistently surpasses the baseline in all languages. This suggests that CLEAR can help enhance semantic representation within individual languages, in addition to improving alignment across languages.

These advantages can be attributed to the formation of a shared embedding space. In Figure~\ref{fig:multi_dist}, we observe that CLEAR demonstrates significantly better language-agnostic alignment for languages and achieves closer semantic proximity between the gold passage and query than InfoNCE. This supports our claim that CLEAR constructs a robust language-agnostic space by narrowing the fundamental distance between language spaces and mapping them into a similar embedding space via passage bridge. CLEAR can be extended to encompass multilingualism, showing broader applicability.

\subsection{Ablation Study}
\label{sec:5.3}
In this section, we analyze the influence of core strategies in CLEAR. In the cross-lingual scenario, we sequentially remove each proposed strategy from CLEAR and evaluate the impact on performance. Table~\ref{tab:ablation} shows that each has a substantial impact on performance and also proves that the three strategies work in synergy for enhancing overall cross-lingual retrieval performance.

\paragraph{Validity of Passage Bridge}
We find that the passage bridge plays a vital role in cross-lingual alignment. In general, when the loss function jointly optimizes both the English training~(\(\mathcal{L}_{\text{NCE}_{\text{en}}}\)) and the cross-lingual training objective~(\(\mathcal{L}_{\text{CL}}\)), the model’s capacity is distributed between these objectives, which may reduce the concentration on cross-lingual alignment during training.

However, the exclusion of the passage bridge via the removal of the English objective term leads to the most notable decrease in performance. We ascribe this phenomenon to the shared passages, which function as bridges that interlink the representation spaces of different languages and facilitate the sharing of semantic information. By positioning target languages closer to English within the embedding space via the passage bridge, CLEAR benefits from cross-lingual alignment.

\paragraph{Importance of Reversal Scheme}
Training with the conventional direction~(where the query serves as the anchor in $\mathcal{L}_{\text{CL}}$) instead of the reversal scheme consistently degrades performance in the majority of cases. This is surprising given that reverse training does not directly align with the retrieval task's objective of finding relevant passages for a query. Also, this approach weakens the synergy with the passage bridge. These findings imply that aligning gold pairs through various perspectives beyond the conventional direction, in conjunction with a passage bridge, enhances the robustness of cross-lingual alignment.

Moreover, CLEAR's efficacy does not simply arise from increased computational demands. Substituting the reversal scheme with conventional direction also maintains all loss terms, leaving the computation amount identical to the case where the reversal scheme is applied. Given this, we can attribute CLEAR’s benefit to the proposed reversal training scheme itself.

\paragraph{KL-Divergence}
The introduction of \(\mathcal{L}_{\text{KL}}\) leads to a slight improvement in overall performance. This gain stems from harmonizing the patterns of similarity observed among English pairs with those between English passages and target language queries within each training batch. By ingraining a more fundamental understanding of cross-lingual semantic equivalence, \(\mathcal{L}_{\text{KL}}\) complements the instance-level alignments fostered by other strategies.

\section{Conclusion}
We propose CLEAR, an innovative approach designed to enhance cross-lingual alignment within the realm of cross-lingual information retrieval. Our reversal training scheme, coupled with several strategies, promotes diverse interaction between English and the target language. Through the experiments on nine languages, CLEAR outperforms the standard approach, demonstrating notable robustness and adaptability in diverse linguistic environments. Furthermore, we highlight the efficacy of CLEAR in extending beyond cross-lingual to multilingual setups, showcasing its utility across broader scenarios. Our study suggests that CLEAR offers a promising avenue for future research and application in global information retrieval systems, which can be directly integrated into existing dense retrieval frameworks. For future work, we plan to explore the expansion of its application to other text embedding tasks beyond retrieval.

\section*{Limitation}
Our study primarily focused on cross-lingual scenarios involving English and other target languages. Although broader coverage could be achieved by considering scenarios where both the passage and query languages are non-English, this was challenging for us due to the limited resources of the parallel dataset, especially at the passage level. Nevertheless, by achieving consistent improvements in cross-lingual retrieval adopted in most previous works, motivated by the practical demand for English resources in real-world applications, we were able to demonstrate strong generalization across a wide range of languages.

\section*{Ethics Statement}
In this research, we utilized only publicly available datasets and models. All data used for training and evaluation were sourced from open-access repositories and applied in accordance with their respective licenses. We strictly adhered to the copyright, licensing terms, and guidelines of the original works and datasets, including those pertaining to language resources and translated data. We confirm that there were no distinct ethical concerns related to the collection, use, or processing of the datasets and resources used in this study.

\section*{Acknowledgments}
This research was supported by Basic Science Research Program through the National Research Foundation of Korea~(NRF) funded by the Ministry of Education~(NRF-2021R1A6A1A03045425). This work was supported by the Commercialization Promotion Agency for R\&D Outcomes~(COMPA) grant funded by the Korea government~(Ministry of Science and ICT)(2710086166). This work was supported by Institute for Information \& communications Technology Promotion~(IITP) grant funded by the Korea government~(MSIT) 
(RS-2024-00398115, Research on the reliability and coherence of outcomes produced by Generative AI). This research was supported by Culture, Sports and Tourism R\&D Program through the korea Creative Content Agency grant funded by the Ministry of Culture, Sports and Toruism in 2024~(Project Name : Development of generative AI-based publishing content analysis and sharing platform technology to respond to changes in the publishing environment. Project Number : : RS-2024-00442061).

\bibliography{custom}

\appendix

\section{Training Details}
\label{appen:hyper}
We leveraged the Pytorch framework~\cite{paszke2019pytorch} and the Sentence-Transformers library\footnote{\url{https://github.com/UKPLab/sentence-transformers}}. For the loss function, we use MultipleNegativesRankingLoss\footnote{\url{https://github.com/UKPLab/sentence-transformers/blob/master/sentence_transformers/losses/MultidpleNegativesRankingLoss.py}} as a baseline, which incorporates positive passages with negative samples~\cite{henderson2017efficient}. We used the cached version of loss provided by sentence-transformers for memory efficiency~\cite{gao2021scaling}.

We conducted all experiments under a uniform setup across all languages and models, employing four NVIDIA A6000 GPUs to perform fine-tuning. For hyper-parameters, we set a maximum sequence length of 512 and utilized a batch size of 64 with mini-batch size 32. The learning rate was established at 5e-5, using a cosine scheduler and a warmup ratio of 0.05 for stable training. We reported our experimental results by adopting the last checkpoint after training all models for only one epoch. The details of embedding models employed in our study are shown in Table~\ref{tab:appen_models}.

\begin{table}[htb!]
\centering
\resizebox{\linewidth}{!}{
\begin{tabular}{cl}
\toprule
Model & Details \\
\midrule

\multirow{2}{*}{bge-m3} & \texttt{BAAI/bge-m3} \\ 
& \cite{chen-etal-2024-m3} \\
\multirow{2}{*}{multilingual-e5} & \texttt{intfloat/multilingual-e5-base} \\ 
& \cite{wang2024multilingual} \\
\multirow{2}{*}{gte-multilingual} & \texttt{Alibaba-NLP/gte-multilingual-base} \\ 
& \cite{zhang-etal-2024-mgte} \\
\multirow{2}{*}{jina-v3} & \texttt{jinaai/jina-embeddings-v3} \\ 
& \cite{sturua2024jina} \\

\bottomrule

\end{tabular}}
\caption{Embedding models details.}
\label{tab:appen_models}
\end{table}

\section{Evaluation Benchmark}
\label{appen:benchmark}
In this work, we leverage multilingual Question Answering~(QA) datasets with parallel constructions, repurposed as retrieval tasks, to systematically assess cross-lingual retrieval performance in all directions. Since these datasets are originally designed for QA, the corresponding passages serve as exact gold labels within the retrieval framework. Consequently, datasets developed for QA are widely adopted for the evaluation of the retriever in the current literature~\cite{enevoldsen2025mmteb,lee2025gemini,zhang2025qwen3}. 

\paragraph{Belebele} Belebele is a high-quality, professionally translated multilingual QA dataset featuring a broad range of language pairs. All translations were conducted by native speakers proficient in English, thereby capturing both contextual meaning and cultural nuances. Owing to these strengths, Belebele offers diverse and realistic multilingual retrieval scenarios, enabling detailed comparative analyses of retrieval models across different languages.

\paragraph{XQuAD} XQuAD is a multilingual QA resource based on SQuAD 1.1~\cite{rajpurkar-etal-2016-squad}, comprising fully parallel question-answer pairs spanning 13 languages, including English. The dataset was translated by professional translators, ensuring strict one-to-one mapping between documents and queries across languages. This rigorous translation approach preserves both linguistic characteristics and semantic content in each target language, rendering XQuAD especially well-suited for evaluating the stability of embedding models with respect to linguistic variation in cross-lingual contexts.

We also considered a wide range of datasets for cross-lingual evaluation. However, most retrieval datasets either have quality issues that affect their reliability or do not align well with the objectives of our study. For example, since Mr.TyDi~\cite{zhang-etal-2021-mr} and MIRACL~\cite{zhang-etal-2023-miracl} are not fully parallel with English, they cannot be used for cross-lingual evaluation. In this regime, we carefully select Belebele and XQuAD as our main evaluation datasets.

\section{XOR-TyDi}
Additionally, to further substantiate our evaluation, we report the cross-lingual evaluation results in the common retrieval task, XOR-TyDi~\cite{asai-etal-2021-xor}. We utilize the XOR-Retrieve task from XOR-TyDi and conduct supplementary experiments on the four languages that overlap with those covered in our study, out of the seven languages included in the benchmark. The evaluation is performed using an English passage and target language queries for both the cross-lingual and multilingual training scenarios, as XOR-TyDi does not provide support for passages in target languages.

\begin{table}[t]
\centering
\renewcommand{\arraystretch}{1.1}
\resizebox{\linewidth}{!}{
\begin{tabular}{lcccccccccc}
\toprule
\multirow{2}{*}{\textbf{Lang}}
    & \multicolumn{2}{c}{\textbf{bge-m3}}
    & \multicolumn{2}{c}{\textbf{multilingual-e5}}
    & \multicolumn{2}{c}{\textbf{gte-multilingual}}
    & \multicolumn{2}{c}{\textbf{jina-v3}} \\
\cmidrule(lr){2-3}
\cmidrule(lr){4-5}
\cmidrule(lr){6-7}
\cmidrule(lr){8-9}
    & InfoNCE & CLEAR & InfoNCE & CLEAR & InfoNCE & CLEAR & InfoNCE & CLEAR \\

\midrule[1.2pt]
\multicolumn{11}{c}{\textit{\textbf{Cross-lingual Scenario}}}\\
\midrule[1.2pt]
ar  & 72.44 & \textbf{72.95} & 64.11 & \textbf{64.99} & 66.22 & \textbf{66.75} & 71.77 & \textbf{71.89} \\
ru  & \textbf{76.93} & 76.77 & 70.84 & \textbf{71.83} & 74.36 & \textbf{74.76} & 76.99 & \textbf{77.59} \\
te  & 68.78 & \textbf{69.66} & 55.56 & \textbf{59.42} & 68.36 & \textbf{69.26} & 67.63 & \textbf{67.83} \\
bn  & 69.96 & \textbf{69.90} & 60.30 & \textbf{62.24} & 63.67 & \textbf{64.30} & 68.59 & \textbf{68.95} \\
\midrule
\textbf{Avg.} & 72.03 & \textbf{72.32} & 62.20 & \textbf{64.12} & 68.15 & \textbf{68.77} & 71.25 & \textbf{71.57} \\

\midrule[1.2pt]
\multicolumn{11}{c}{\textit{\textbf{Multilingual Expansion}}}\\
\midrule[1.2pt]
ar  & 72.41 & \textbf{72.96} & 64.19 & \textbf{65.46} & 65.71 & \textbf{66.43} & 71.02 & \textbf{71.76} \\
ru  & \textbf{76.69} & 76.62 & 70.01 & \textbf{71.52} & 74.29 & \textbf{74.83} & \textbf{77.07} & 77.05 \\
te  & 69.61 & \textbf{70.42} & 54.96 & \textbf{57.63} & 68.33 & \textbf{68.98} & 67.45 & \textbf{67.98} \\
bn  & 70.11 & \textbf{70.43} & 59.78 & \textbf{60.70} & 63.32 & \textbf{64.14} & 68.38 & \textbf{68.71} \\
\midrule
\textbf{Avg.} & 72.21 & \textbf{72.61} & 62.23 & \textbf{63.83} & 67.41 & \textbf{68.09} & 70.98 & \textbf{71.38} \\
\bottomrule
\end{tabular}}
\caption{Cross-lingual evaluation results in XOR-TyDi. In the cross-lingual scenario, the results are from models fine-tuned individually for each language. In the multilingual expansion, the model is trained considering all languages equally, as in our main experiments.}
\label{tab:xorqa}
\end{table}

As can be seen in the Table~\ref{tab:xorqa}, the result is in line with our main experimental result. This emphasizes again that CLEAR is indeed an effective training approach for cross-lingual retrieval tasks.


\section{Training Example}
\label{appen:train_ex}
Compared to the standard InfoNCE, CLEAR requires translated queries that are parallel to the English query for training. CLEAR does not require target language passages. Thus, $Q_{en}$ is used as the anchor to increase similarity with $P^{+}_{en}$ and decrease similarity with $P^{-}_{en}$ in $L_{NCE_{en}}$. In $L_{CL}$, $P^{+}_{en}$ serves as the anchor, with $Q_{\ell}$ treated as a positive and $Q^{-}_{\ell}$ as a hard negative sample in contrastive learning. 
An example of inputs for the training is shown in Table~\ref{tab:train_ex}.

\begin{table}[h!]
\centering
\renewcommand{\arraystretch}{1.1}
\resizebox{\linewidth}{!}{
\begin{tabular}{p{1cm}p{11cm}}
\toprule
$Q_{en}$ & When was Winehouse quoted about the album release? \\
\midrule
$P^{+}_{en}$ & Winehouse and Ronson contributed a cover of Lesley Gore's ``It's My Party'' to the Quincy Jones tribute album Q Soul Bossa Nostra released 9 November 2010 ... In July 2010, Winehouse was quoted as saying her next album would be released no later than January 2011, saying ``It's going to be very much the same as my second album, where there's a lot of jukebox stuff and songs that are... just jukebox, really.'' \\
\midrule
$P^{-}_{en}$ & On the other hand, Arabic is divided into over 27 dialects. Almost every Arab state has at least one local dialect of its own ... \\
\midrule
\multirow{4}{*}{$Q_{\ell}$} & zh: 懷恩豪斯何時提到這張專輯的發行? \\
 & de: Wann hat Winehouse die Veröffentlichung dieses Albums erwähnt? \\
 & ru: Когда Уайнхаус упомянула о выпуске этого альбома?\\
 & es: ¿Cuándo mencionó Winehouse el lanzamiento de este álbum?\\
\midrule
\multirow{4}{*}{$Q^{-}_{\ell}$} & zh: 超級女孩電視劇什麼時候首次播出? \\
 & de: Was tut die Polizei, um Sarah zu beschützen? \\
 & ru: Когда родился Жак-Луи Давид? \\
 & es: ¿Cuántos huevos pueden poner las aves en las colinas de Sri Lanka? \\
\bottomrule
\end{tabular}}
\caption{Example of inputs in the training.}
\label{tab:train_ex}
\end{table}

\section{Robustness Analysis on Translation Quality}
\begin{table}[t]
\centering
\renewcommand{\arraystretch}{1.1}
\resizebox{\linewidth}{!}{
\begin{tabular}{llcccc}
\toprule
\textbf{Model} & \textbf{Lang} & \multicolumn{2}{c}{\textbf{English-Lang}} & \multicolumn{2}{c}{\textbf{Lang-English}} \\
\cmidrule(lr){3-4}
\cmidrule(lr){5-6}
 &  & InfoNCE & CLEAR & InfoNCE & CLEAR \\
\toprule
\multirow{8}{*}{bge-m3} & ar & 88.66 & \textbf{89.23} & 90.87 & \textbf{91.17} \\
 & zh & 91.86 & \textbf{92.40} & 91.63 & \textbf{92.10} \\
 & es & 92.66 & \textbf{93.17} & 93.07 & \textbf{93.44} \\
 & de & 93.77 & \textbf{94.30} & 93.99 & \textbf{94.66} \\
 & ru & 93.72 & \textbf{94.17} & 93.42 & \textbf{93.54} \\
 & hi & 89.15 & \textbf{89.91} & 93.33 & \textbf{93.47} \\
 & vi & 92.79 & \textbf{93.34} & 92.77 & \textbf{93.35} \\
 & bn & 89.49 & \textbf{90.43} & 92.24 & \textbf{93.04} \\
\midrule
\multirow{8}{*}{multilingual-gte} & ar & 86.49 & \textbf{88.10} & 90.40 & \textbf{91.62} \\
 & zh & 92.34 & \textbf{92.83} & 91.99 & \textbf{92.70} \\
 & es & 93.05 & \textbf{93.52} & 93.41 & \textbf{93.93} \\
 & de & 92.78 & \textbf{93.42} & 93.53 & \textbf{94.00} \\
 & ru & 92.34 & \textbf{92.88} & 93.22 & \textbf{93.78} \\
 & hi & 89.17 & \textbf{89.92} & 92.66 & \textbf{93.22} \\
 & vi & 91.46 & \textbf{92.11} & 92.99 & \textbf{93.55} \\
 & bn & 85.29 & \textbf{86.28} & 90.98 & \textbf{91.91} \\
\bottomrule
\end{tabular}
}
\caption{Cross-lingual evaluation result in Belebele trained with m2m100 translated dataset.}
\label{tab:train_trans}
\end{table}
To evaluate the robustness of our approach against noise that may influence the results because of machine-translated training data, we conduct further experiments using another translation model, m2m100~\cite{fan2021beyond}\footnote{\url{https://huggingface.co/facebook/m2m100_1.2B}}, within the same pipeline (except Telugu language, since m2m100 does not support). As shown in Table~\ref{tab:train_trans}, CLEAR consistently yields performance improvements at the same level of data quality, aligning with our main result.

\begin{figure*}[h!]
\centering
\subcaptionbox{English - Lang}{\includegraphics[width=\textwidth]{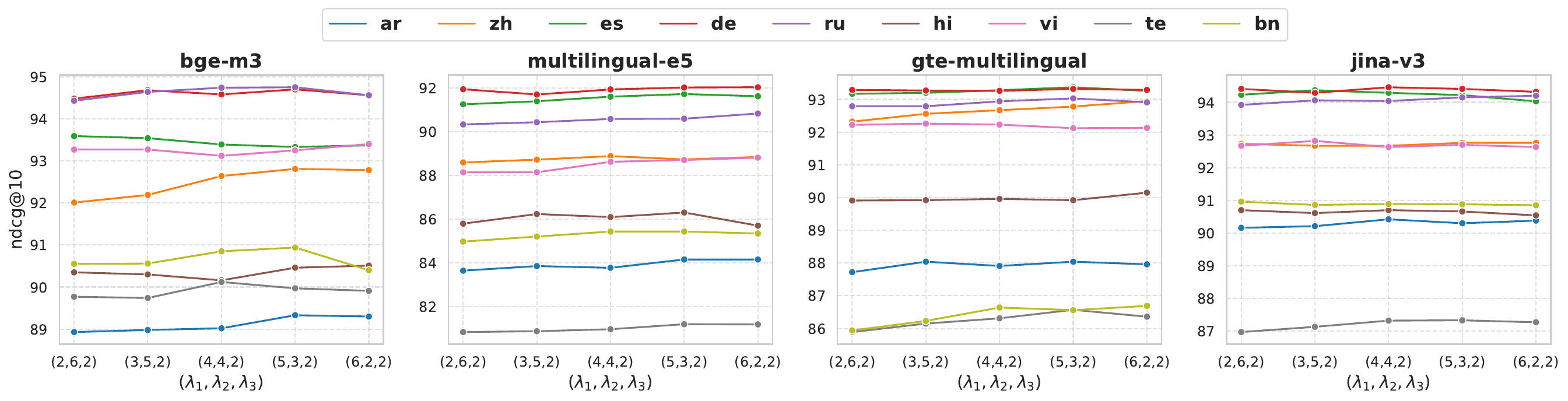}}
\vfill
\subcaptionbox{Lang - English}{\includegraphics[width=\textwidth]{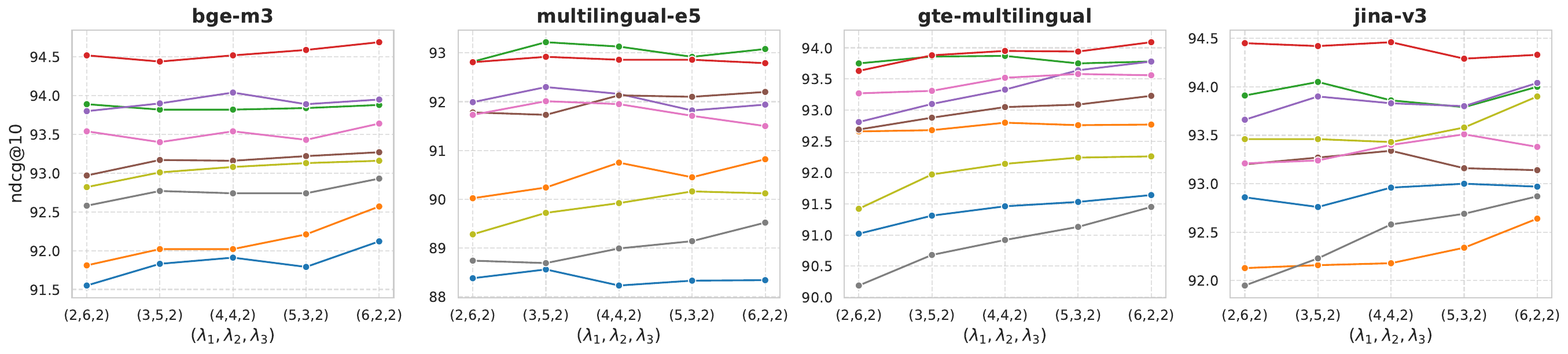}}
\vfill
\caption{Performance variation depending on the loss component weights ($\lambda_{1}$, $\lambda_{2}$, $\lambda_{3}$) across multiple languages.}
\label{fig:loss_param}
\end{figure*}

From the perspective of robustness, we observe no significant performance difference between models trained on m2m100-translated data and those trained on NLLB-translated data. Since CLEAR requires only translated queries rather than complex passages, potential translation errors are likely minimized. Our result demonstrates that CLEAR can be an essentially robust approach with respect to variations in translation quality, thereby enhancing its practical applicability in real-world scenarios where human translation is costly.

\section{Loss Parameter Sensitivity Analysis}
\label{appen:loss_param}

We report the nDCG@10 on Belebele for the settings of English passage-target language query and target language passage-English query in Figure~\ref{fig:loss_param}. While exhaustively exploring every possible combination of values for all three components is computationally prohibitive, we focus our analysis on the components hypothesized to exert the most significant influence on CLEAR’s core objectives: $\lambda_{1}$, which controls the English NCE term, and $\lambda_{2}$, which governs the cross-lingual reversal loss.

Overall, the results indicate that variations in the $\lambda$ values do not cause substantial performance fluctuations, suggesting that the proposed method is relatively robust to the choice of loss-weight parameters. In the case of English passages with target-language queries (Figure~\ref{fig:loss_param}~(a)), the combinations (4, 4, 2) and (5, 3, 2) yield the best average performance across all models. However, we observe that the Lang-English setup benefits from a stronger alignment of English representations (Figure~\ref{fig:loss_param}~(b)). This pattern highlights that our bridging strategy, which adopts the English representation as a semantic anchor between languages, plays a crucial role in enhancing the passage representation capability of different languages. Moreover, these results suggest that there remains room for further improvement by refining the loss weight configuration, particularly in the Lang-English setup.

\section{Extended Evaluation Results}
\label{appen:all_results}
Among the selected nine languages, Belebele encompasses all languages, and XQuAD includes only five. We report the extended evaluation results in this section. Table~\ref{tab:belebele_full},~\ref{tab:xquad_full} presents the results for each target language under the cross-lingual setup, and Table~\ref{tab:belebele_full_multi},~\ref{tab:xquad_full_multi} illustrates the performance in the multilingual training setup.

\begin{table*}[htb!]
\centering
\scriptsize
\resizebox{\textwidth}{!}{%
\begin{tabular}{cccccccccccccc}

\toprule
\multirow{3}{*}{\textbf{Model}} & \multirow{3}{*}{\textbf{Language}} 
& \multicolumn{6}{c}{\textbf{English-Lang}} & \multicolumn{6}{c}{\textbf{Lang-English}} \\
\cmidrule(lr){3-8}\cmidrule(lr){9-14}
& 
    & \multicolumn{3}{c}{\textbf{nDCG@5}} & \multicolumn{3}{c}{\textbf{nDCG@10}}
    & \multicolumn{3}{c}{\textbf{nDCG@5}} & \multicolumn{3}{c}{\textbf{nDCG@10}} \\
    \cmidrule(lr){3-5}\cmidrule(lr){6-8}\cmidrule(lr){9-11}\cmidrule(lr){12-14}
& 
    & \textbf{Base} & \textbf{InfoNCE} & \textbf{CLEAR}& \textbf{Base} & \textbf{InfoNCE} & \textbf{CLEAR}& \textbf{Base} & \textbf{InfoNCE} & \textbf{CLEAR} & \textbf{Base} & \textbf{InfoNCE} & \textbf{CLEAR} \\
\midrule

\multirow{9}{*}{bge-m3}
  & ar & 86.62 & 88.22 & \textbf{88.29}
       & 87.77 & \textbf{89.06} & 89.02
       & 86.42 & 90.66 & \textbf{90.93}
       & 87.63 & 91.61 & \textbf{91.91} \\
  & zh & 89.70 & 90.93 & \textbf{92.14}
       & 90.35 & 91.69 & \textbf{92.64}
       & 89.08 & 90.97 & \textbf{91.51}
       & 89.81 & 91.63 & \textbf{92.02} \\
  & es & 91.48 & 92.61 & \textbf{93.05}
       & 91.99 & 92.81 & \textbf{93.39}
       & 91.47 & 92.99 & \textbf{93.59}
       & 92.05 & 93.50 & \textbf{93.82} \\
  & de & 92.20 & 93.62 & \textbf{94.31}
       & 92.81 & 93.98 & \textbf{94.58}
       & 92.47 & 93.87 & \textbf{94.30}
       & 92.90 & 94.30 & \textbf{94.52} \\
  & ru & 92.61 & 93.28 & \textbf{94.30}
       & 92.92 & 93.67 & \textbf{94.74}
       & 91.18 & 92.85 & \textbf{93.61}
       & 91.69 & 93.55 & \textbf{94.04} \\
  & hi & 85.78 & 88.15 & \textbf{89.72}
       & 86.90 & 89.14 & \textbf{90.16}
       & 89.57 & 92.35 & \textbf{92.62}
       & 90.39 & 92.86 & \textbf{93.16} \\
  & vi & 90.02 & 92.26 & \textbf{92.79}
       & 90.70 & 92.58 & \textbf{93.12}
       & 89.73 & 92.53 & \textbf{93.10}
       & 90.44 & 93.04 & \textbf{93.54} \\
  & te & 85.98 & 88.33 & \textbf{89.49}
       & 87.12 & 89.07 & \textbf{90.12}
       & 88.27 & 91.41 & \textbf{92.25}
       & 89.19 & 92.08 & \textbf{92.74} \\
  & bn & 86.49 & 88.95 & \textbf{90.12}
       & 87.58 & 89.79 & \textbf{90.85}
       & 89.18 & 91.65 & \textbf{92.71}
       & 90.17 & 92.37 & \textbf{93.08} \\
\midrule
\multirow{9}{*}{multilingual-e5}
  & ar & 76.23 & 80.68 & \textbf{82.79}
       & 77.65 & 82.27 & \textbf{83.78}
       & 82.51 & 87.24 & \textbf{87.39}
       & 83.56 & 87.96 & \textbf{88.23} \\
  & zh & 80.04 & 85.82 & \textbf{87.83}
       & 81.16 & 87.00 & \textbf{88.89}
       & 84.43 & 89.15 & \textbf{89.87}
       & 85.49 & 90.02 & \textbf{90.75} \\
  & es & 88.53 & 89.01 & \textbf{91.08}
       & 89.46 & 89.84 & \textbf{91.61}
       & 91.92 & 91.79 & \textbf{92.85}
       & 92.43 & 92.15 & \textbf{93.13} \\
  & de & 89.96 & 90.26 & \textbf{91.44}
       & 90.77 & 90.97 & \textbf{91.94}
       & 91.53 & 91.78 & \textbf{92.17}
       & 92.16 & 92.68 & \textbf{92.86} \\
  & ru & 86.48 & 88.82 & \textbf{89.79}
       & 87.29 & 89.60 & \textbf{90.59}
       & 87.73 & 90.72 & \textbf{91.48}
       & 88.98 & 91.55 & \textbf{92.16} \\
  & hi & 79.46 & 81.82 & \textbf{84.88}
       & 81.05 & 83.11 & \textbf{86.10}
       & 87.65 & 90.34 & \textbf{91.52}
       & 88.44 & 91.08 & \textbf{92.13} \\
  & vi & 85.05 & 86.18 & \textbf{87.76}
       & 86.03 & 87.11 & \textbf{88.63}
       & 87.52 & 91.21 & \textbf{91.44}
       & 88.58 & 91.81 & \textbf{91.95} \\
  & te & 67.87 & 75.08 & \textbf{79.40}
       & 69.90 & 77.14 & \textbf{80.97}
       & 83.57 & 86.90 & \textbf{88.36}
       & 84.88 & 88.05 & \textbf{88.99} \\
  & bn & 73.83 & 80.34 & \textbf{84.62}
       & 75.55 & 81.81 & \textbf{85.44}
       & 82.95 & 88.22 & \textbf{89.22}
       & 84.31 & 88.91 & \textbf{89.92} \\
\midrule
\multirow{9}{*}{gte-multilingual}
  & ar & 82.60 & 86.27 & \textbf{87.32}
       & 83.81 & 86.92 & \textbf{87.91}
       & 86.92 & 90.58 & \textbf{91.06}
       & 87.99 & 90.97 & \textbf{91.46} \\
  & zh & 89.13 & 91.82 & \textbf{92.11}
       & 89.86 & 92.30 & \textbf{92.67}
       & 90.92 & 91.50 & \textbf{92.14}
       & 91.51 & 92.05 & \textbf{92.80} \\
  & es & 90.78 & 92.09 & \textbf{92.85}
       & 91.71 & 92.86 & \textbf{93.28}
       & 89.73 & 93.08 & \textbf{93.40}
       & 90.20 & 93.55 & \textbf{93.87} \\
  & de & 90.56 & 92.51 & \textbf{93.04}
       & 91.21 & 92.92 & \textbf{93.26}
       & 88.63 & 93.11 & \textbf{93.63}
       & 89.42 & 93.45 & \textbf{93.95} \\
  & ru & 90.27 & 91.95 & \textbf{92.50}
       & 91.11 & 92.46 & \textbf{92.94}
       & 88.57 & 92.28 & \textbf{93.04}
       & 89.56 & 92.69 & \textbf{93.33} \\
  & hi & 86.49 & 88.60 & \textbf{89.39}
       & 87.51 & 89.34 & \textbf{89.96}
       & 88.96 & 91.88 & \textbf{92.60}
       & 89.55 & 92.45 & \textbf{93.05} \\
  & vi & 88.68 & 91.01 & \textbf{91.68}
       & 89.37 & 91.59 & \textbf{92.23}
       & 89.88 & 92.65 & \textbf{93.05}
       & 90.48 & 93.14 & \textbf{93.52} \\
  & te & 79.13 & 83.53 & \textbf{85.00}
       & 80.70 & 84.77 & \textbf{86.31}
       & 87.49 & 89.18 & \textbf{90.41}
       & 88.46 & 89.96 & \textbf{90.92} \\
  & bn & 80.76 & 83.89 & \textbf{85.65}
       & 82.13 & 85.18 & \textbf{86.64}
       & 87.09 & 90.47 & \textbf{91.68}
       & 87.81 & 91.03 & \textbf{92.14} \\
\midrule
\multirow{9}{*}{jina-v3}
  & ar & 85.85 & 88.43 & \textbf{89.52}
       & 86.84 & 89.37 & \textbf{90.42}
       & 89.12 & 91.63 & \textbf{92.31}
       & 89.86 & 92.45 & \textbf{92.96} \\
  & zh & 88.93 & 91.24 & \textbf{92.08}
       & 89.46 & 91.90 & \textbf{92.67}
       & 88.88 & 91.06 & \textbf{91.65}
       & 89.64 & 91.74 & \textbf{92.18} \\
  & es & 90.88 & 93.14 & \textbf{94.00}
       & 91.40 & 93.57 & \textbf{94.29}
       & 91.95 & 93.46 & \textbf{93.59}
       & 92.64 & \textbf{93.88} & 93.86 \\
  & de & 91.20 & 93.42 & \textbf{93.91}
       & 91.75 & 93.88 & \textbf{94.46}
       & 91.68 & 93.61 & \textbf{94.05}
       & 92.39 & 94.12 & \textbf{94.46} \\
  & ru & 91.53 & 93.39 & \textbf{93.56}
       & 91.85 & 93.77 & \textbf{94.04}
       & 90.82 & 92.98 & \textbf{93.46}
       & 91.66 & 93.46 & \textbf{93.83} \\
  & hi & 86.94 & 89.21 & \textbf{90.34}
       & 87.74 & 89.90 & \textbf{90.70}
       & 90.77 & 92.52 & \textbf{93.01}
       & 91.50 & 93.02 & \textbf{93.34} \\
  & vi & 89.39 & 91.76 & \textbf{92.01}
       & 90.26 & \textbf{92.68} & 92.63
       & 90.10 & 92.90 & \textbf{93.05}
       & 90.98 & 93.12 & \textbf{93.40} \\
  & te & 81.49 & 84.35 & \textbf{86.39}
       & 83.02 & 85.57 & \textbf{87.32}
       & 88.15 & 91.33 & \textbf{92.10}
       & 88.99 & 91.87 & \textbf{92.58} \\
  & bn & 85.45 & 88.27 & \textbf{89.94}
       & 86.56 & 89.22 & \textbf{90.89}
       & 90.48 & 93.05 & \textbf{92.95}
       & 91.14 & 93.52 & \textbf{93.43} \\
\bottomrule
\end{tabular}%
}
\caption{Results on all languages under cross-lingual scenario in Belebele.}
\label{tab:belebele_full}
\end{table*}

\begin{table*}[htb!]
\centering
\scriptsize
\resizebox{\textwidth}{!}{%
\begin{tabular}{cccccccccccccc}
\toprule
\multirow{4}{*}{\textbf{Model}} & \multirow{4}{*}{\textbf{Language}}
  & \multicolumn{6}{c}{\textbf{English-Lang}}
  & \multicolumn{6}{c}{\textbf{Lang-English}} \\
\cmidrule(lr){3-8}\cmidrule(lr){9-14}
&  & \multicolumn{3}{c}{\textbf{nDCG@5}} & \multicolumn{3}{c}{\textbf{nDCG@10}}
   & \multicolumn{3}{c}{\textbf{nDCG@5}} & \multicolumn{3}{c}{\textbf{nDCG@10}} \\
\cmidrule(lr){3-5}\cmidrule(lr){6-8}\cmidrule(lr){9-11}\cmidrule(lr){12-14}
& 
  & \textbf{Base} & \textbf{InfoNCE} & \textbf{CLEAR}
  & \textbf{Base} & \textbf{InfoNCE} & \textbf{CLEAR}
  & \textbf{Base} & \textbf{InfoNCE} & \textbf{CLEAR}
  & \textbf{Base} & \textbf{InfoNCE} & \textbf{CLEAR} \\
\midrule
\multirow{5}{*}{bge-m3}
  & ar  & 91.84 & 92.69 & \textbf{93.24}
        & 92.24 & 93.08 & \textbf{93.50}
        & 91.91 & 94.03 & \textbf{94.44}
        & 92.38 & 94.41 & \textbf{94.71} \\
  & zh  & 93.85 & 93.66 & \textbf{94.34}
        & 94.04 & 94.03 & \textbf{94.68}
        & 92.74 & 93.80 & \textbf{94.50}
        & 93.18 & 94.21 & \textbf{94.82} \\
  & es  & \textbf{95.98} & 95.86 & 95.97
        & 96.14 & 95.97 & \textbf{96.14}
        & 95.80 & \textbf{96.20} & 96.09
        & 95.96 & \textbf{96.42} & 96.34 \\
  & de  & 95.29 & 95.64 & \textbf{95.91}
        & 95.52 & 95.75 & \textbf{96.04}
        & 95.89 & 95.53 & \textbf{95.91}
        & \textbf{96.17} & 95.67 & 95.93 \\
  & ru  & 95.29 & 94.91 & \textbf{95.67}
        & 95.58 & 95.13 & \textbf{95.83}
        & 94.17 & 94.73 & \textbf{94.83}
        & 94.57 & 94.98 & \textbf{95.16} \\
\midrule
\multirow{5}{*}{multilingual-e5}
  & ar  & 86.36 & 86.88 & \textbf{89.31}
        & 87.29 & 87.58 & \textbf{89.83}
        & 90.82 & 91.27 & \textbf{92.36}
        & 91.22 & 91.69 & \textbf{92.61} \\
  & zh  & 88.65 & 90.22 & \textbf{91.27}
        & 89.60 & 90.64 & \textbf{91.71}
        & 90.47 & 92.18 & \textbf{93.36}
        & 91.02 & 92.44 & \textbf{93.44} \\
  & es  & \textbf{96.01} & 93.06 & 93.75
        & \textbf{96.15} & 93.39 & 94.00
        & \textbf{96.25} & 94.18 & 94.39
        & \textbf{96.41} & 94.42 & 94.62 \\
  & de  & \textbf{94.84} & 92.01 & 92.49
        & \textbf{95.18} & 92.39 & 92.93
        & \textbf{95.41} & 93.45 & 93.49
        & \textbf{95.63} & 93.76 & 93.79 \\
  & ru  & \textbf{92.50} & 91.42 & 92.09
        & \textbf{93.06} & 91.90 & 92.54
        & \textbf{92.88} & 92.13 & 92.62
        & \textbf{93.22} & 92.54 & 93.11 \\
\midrule
\multirow{5}{*}{gte-multilingual}
  & ar  & 86.65 & 87.99 & \textbf{89.27}
        & 87.44 & 88.59 & \textbf{89.78}
        & 91.29 & 92.83 & \textbf{93.34}
        & 91.89 & 93.26 & \textbf{93.79} \\
  & zh  & 93.70 & 94.08 & \textbf{94.60}
        & 93.98 & 94.27 & \textbf{94.87}
        & 91.61 & 93.43 & \textbf{93.74}
        & 92.07 & 93.67 & \textbf{94.05} \\
  & es  & 95.85 & 95.70 & \textbf{96.29}
        & 96.14 & 95.88 & \textbf{96.45}
        & 95.48 & 96.36 & \textbf{96.55}
        & 95.70 & 96.53 & \textbf{96.66} \\
  & de  & 94.81 & 94.37 & \textbf{94.97}
        & 95.11 & 94.59 & \textbf{95.37}
        & 94.47 & 95.42 & \textbf{95.97}
        & 94.75 & 95.73 & \textbf{96.11} \\
  & ru  & \textbf{94.23} & 93.58 & 94.22
        & 94.38 & 94.06 & \textbf{94.57}
        & 93.82 & 94.52 & \textbf{94.85}
        & 94.24 & 94.88 & \textbf{95.21} \\
\midrule
\multirow{5}{*}{jina-v3}
  & ar  & 90.01 & 92.92 & \textbf{93.68}
        & 90.58 & 93.25 & \textbf{93.90}
        & 92.83 & 94.82 & \textbf{95.11}
        & 93.21 & 95.02 & \textbf{95.19} \\
  & zh  & 92.32 & 94.49 & \textbf{94.77}
        & 92.65 & 94.72 & \textbf{95.05}
        & 92.37 & 95.11 & \textbf{95.59}
        & 92.79 & 95.36 & \textbf{95.75} \\
  & es  & 94.37 & \textbf{95.76} & 95.75
        & 94.76 & \textbf{96.00} & 95.92
        & 95.91 & 96.75 & \textbf{96.88}
        & 96.12 & 96.90 & \textbf{97.02} \\
  & de  & 94.52 & 96.02 & \textbf{96.23}
        & 94.69 & 96.16 & \textbf{96.29}
        & 95.15 & 96.35 & \textbf{96.61}
        & 95.30 & 96.44 & \textbf{96.69} \\
  & ru  & 93.77 & 95.76 & \textbf{95.92}
        & 93.97 & 95.88 & \textbf{96.06}
        & 94.06 & 95.34 & \textbf{95.57}
        & 94.38 & 95.53 & \textbf{95.82} \\
\bottomrule
\end{tabular}%
}
\caption{Results on all languages under cross-lingual scenario in XQuAD.}
\label{tab:xquad_full}
\end{table*}

\begin{table*}[t]
\centering
\scriptsize
\resizebox{\textwidth}{!}{%
\begin{tabular}{cccccccccccccc}
\toprule
\multirow{3}{*}{\textbf{Model}} & \multirow{3}{*}{\textbf{Language}} 
  & \multicolumn{6}{c}{\textbf{English-Lang}} 
  & \multicolumn{6}{c}{\textbf{Lang-English}} \\
\cmidrule(lr){3-8}\cmidrule(lr){9-14}
& 
  & \multicolumn{3}{c}{\textbf{nDCG@5}} & \multicolumn{3}{c}{\textbf{nDCG@10}}
  & \multicolumn{3}{c}{\textbf{nDCG@5}} & \multicolumn{3}{c}{\textbf{nDCG@10}} \\
\cmidrule(lr){3-5}\cmidrule(lr){6-8}\cmidrule(lr){9-11}\cmidrule(lr){12-14}
& 
  & \textbf{Base} & \textbf{InfoNCE} & \textbf{CLEAR}
  & \textbf{Base} & \textbf{InfoNCE} & \textbf{CLEAR}
  & \textbf{Base} & \textbf{InfoNCE} & \textbf{CLEAR}
  & \textbf{Base} & \textbf{InfoNCE} & \textbf{CLEAR} \\
\midrule

\multirow{9}{*}{bge-m3}
  & ar & 86.62 & 88.85 & \textbf{89.31}  & 87.77 & 89.56 & \textbf{89.97}  
       & 86.42 & 90.35 & \textbf{91.01}  & 87.63 & 91.30 & \textbf{91.90} \\
  & zh & 89.70 & 92.31 & \textbf{92.66}  & 90.35 & 92.92 & \textbf{93.10}  
       & 89.08 & 91.55 & \textbf{92.23}  & 89.81 & 92.10 & \textbf{92.67} \\
  & es & 91.48 & 92.03 & \textbf{93.22}  & 91.99 & 92.66 & \textbf{93.44}  
       & 91.47 & 92.60 & \textbf{93.29}  & 92.05 & 93.26 & \textbf{93.81} \\
  & de & 92.20 & 93.44 & \textbf{94.40}  & 92.81 & 93.77 & \textbf{94.59}  
       & 92.47 & 93.72 & \textbf{93.98}  & 92.90 & 94.11 & \textbf{94.53} \\
  & ru & 92.61 & 93.43 & \textbf{94.03}  & 92.92 & 93.94 & \textbf{94.43}  
       & 91.18 & 92.72 & \textbf{93.44}  & 91.69 & 93.14 & \textbf{93.90} \\
  & hi & 85.78 & 88.64 & \textbf{89.48}  & 86.90 & 89.39 & \textbf{90.27}  
       & 89.57 & 92.52 & \textbf{93.15}  & 90.39 & 93.06 & \textbf{93.61} \\
  & vi & 90.02 & 92.39 & \textbf{93.03}  & 90.70 & 92.58 & \textbf{93.12}  
       & 89.73 & 92.54 & \textbf{93.25}  & 90.44 & 93.04 & \textbf{93.54} \\
  & te & 85.98 & 88.26 & \textbf{89.09}  & 92.58 & 92.94 & \textbf{93.50}  
       & 88.27 & 91.78 & \textbf{92.63}  & 93.04 & 93.08 & \textbf{93.62} \\
  & bn & 86.49 & 89.03 & \textbf{89.81}  & 87.58 & 90.00 & \textbf{90.68}  
       & 89.18 & 91.67 & \textbf{92.82}  & 90.17 & 92.38 & \textbf{93.16} \\
\midrule

\multirow{9}{*}{multilingual-e5}
  & ar & 76.23 & 81.37 & \textbf{83.98}  & 77.65 & 82.58 & \textbf{85.12}  
       & 82.51 & 87.19 & \textbf{88.17}  & 83.56 & 88.08 & \textbf{88.89} \\
  & zh & 80.04 & 85.70 & \textbf{88.10}  & 81.16 & 86.84 & \textbf{88.85}  
       & 84.43 & 89.18 & \textbf{89.98}  & 85.49 & 90.02 & \textbf{90.78} \\
  & es & 88.53 & 88.42 & \textbf{91.50}  & 89.46 & 89.46 & \textbf{91.98}  
       & 91.92 & 90.84 & \textbf{92.74}  & 92.43 & 91.58 & \textbf{93.14} \\
  & de & 89.96 & 89.23 & \textbf{91.32}  & 90.77 & 89.99 & \textbf{91.87}  
       & 91.53 & 91.67 & \textbf{92.97}  & 92.16 & 92.29 & \textbf{93.47} \\
  & ru & 86.48 & 88.68 & \textbf{90.91}  & 87.29 & 89.16 & \textbf{91.68}  
       & 87.73 & 90.64 & \textbf{91.63}  & 88.98 & 91.39 & \textbf{92.25} \\
  & hi & 79.46 & 82.08 & \textbf{84.83}  & 81.05 & 83.10 & \textbf{85.94}  
       & 87.65 & 90.31 & \textbf{91.81}  & 88.44 & 90.97 & \textbf{92.39} \\
  & vi & 85.05 & 85.46 & \textbf{88.31}  & 86.03 & 86.69 & \textbf{88.85}  
       & 87.52 & 89.79 & \textbf{90.95}  & 84.88 & 90.44 & \textbf{91.47} \\
  & te & 67.87 & 74.88 & \textbf{78.65}  & 69.90 & 76.48 & \textbf{80.22}  
       & 83.57 & 87.61 & \textbf{88.96}  & 88.58 & 88.63 & \textbf{89.50} \\
  & bn & 73.83 & 80.33 & \textbf{83.62}  & 75.55 & 81.85 & \textbf{84.51}  
       & 82.95 & 88.17 & \textbf{90.17}  & 84.31 & 88.93 & \textbf{90.68} \\
\midrule

\multirow{9}{*}{gte-multilingual}
  & ar & 82.60 & 85.71 & \textbf{87.33}  & 83.81 & 86.38 & \textbf{87.77}  
       & 86.92 & 90.06 & \textbf{90.87}  & 87.99 & 90.75 & \textbf{91.44} \\
  & zh & 89.13 & 91.20 & \textbf{92.19}  & 89.86 & 91.92 & \textbf{92.70}  
       & 90.92 & 91.46 & \textbf{91.97}  & 91.51 & 92.11 & \textbf{92.52} \\
  & es & 90.78 & 91.49 & \textbf{92.64}  & 91.71 & 92.44 & \textbf{93.29}  
       & 89.73 & 92.63 & \textbf{93.21}  & 90.20 & 93.10 & \textbf{93.53} \\
  & de & 90.56 & 91.65 & \textbf{92.60}  & 91.21 & 92.12 & \textbf{92.92}  
       & 88.63 & 92.87 & \textbf{93.22}  & 89.42 & 93.20 & \textbf{93.55} \\
  & ru & 90.27 & 91.11 & \textbf{92.13}  & 91.11 & 91.80 & \textbf{92.56}  
       & 88.57 & 92.15 & \textbf{92.83}  & 89.56 & 92.71 & \textbf{93.28} \\
  & hi & 86.49 & 88.46 & \textbf{89.59}  & 87.51 & 89.13 & \textbf{90.13}  
       & 88.96 & 91.99 & \textbf{92.45}  & 89.55 & 92.53 & \textbf{92.78} \\
  & vi & 88.68 & 90.58 & \textbf{91.20}  & 89.37 & 91.26 & \textbf{91.82}  
       & 89.88 & 92.47 & \textbf{92.98}  & 90.48 & 92.89 & \textbf{93.23} \\
  & te & 79.13 & 82.51 & \textbf{84.03}  & 80.70 & 83.53 & \textbf{85.19}  
       & 87.49 & 89.30 & \textbf{90.32}  & 90.48 & 89.91 & \textbf{91.03} \\
  & bn & 80.76 & 83.33 & \textbf{84.78}  & 82.13 & 84.42 & \textbf{85.70}  
       & 87.09 & 90.42 & \textbf{91.50}  & 87.81 & 91.08 & \textbf{92.02} \\
\midrule

\multirow{9}{*}{jina-v3}
  & ar & 85.85 & 88.22 & \textbf{89.42}  & 86.84 & 88.88 & \textbf{90.06}  
       & 89.12 & 91.84 & \textbf{92.39}  & 89.86 & 92.37 & \textbf{92.98} \\
  & zh & 88.93 & 91.77 & \textbf{92.37}  & 89.46 & 92.54 & \textbf{93.00}  
       & 88.88 & 91.48 & \textbf{91.91}  & 89.64 & 92.12 & \textbf{92.47} \\
  & es & 90.88 & 92.94 & \textbf{93.71}  & 91.40 & 93.23 & \textbf{94.01}  
       & 91.95 & 93.19 & \textbf{93.62}  & 92.64 & 93.61 & \textbf{93.98} \\
  & de & 91.20 & 93.43 & \textbf{93.82}  & 91.75 & 93.86 & \textbf{94.31}  
       & 91.68 & 93.57 & \textbf{93.94}  & 92.39 & 94.04 & \textbf{94.27} \\
  & ru & 91.53 & 93.59 & \textbf{93.87}  & 91.85 & 93.90 & \textbf{94.24}  
       & 90.82 & \textbf{93.26} & 93.19  & 91.66 & \textbf{93.78} & 93.48 \\
  & hi & 86.94 & 88.97 & \textbf{90.08}  & 87.74 & 89.59 & \textbf{90.51}  
       & 90.77 & 92.53 & \textbf{92.60}  & 91.50 & 92.96 & \textbf{93.08} \\
  & vi & 89.39 & 91.97 & \textbf{92.35}  & 90.26 & 92.84 & \textbf{93.02}  
       & 90.10 & 92.77 & \textbf{92.93}  & 90.98 & 93.09 & \textbf{93.26} \\
  & te & 81.49 & 83.37 & \textbf{85.19}  & 90.26 & 92.84 & \textbf{93.02}  
       & 88.15 & 91.40 & \textbf{91.88}  & 88.99 & 93.09 & \textbf{93.26} \\
  & bn & 85.45 & 88.04 & \textbf{89.79}  & 86.56 & 88.96 & \textbf{90.50}  
       & 90.48 & 92.91 & \textbf{93.24}  & 91.14 & 93.27 & \textbf{93.76} \\
\bottomrule
\end{tabular}%
}
\caption{Results on all languages under the multilingual training setup in Belebele.}
\label{tab:belebele_full_multi}
\end{table*}

\begin{table*}[t]
\centering
\scriptsize
\resizebox{\textwidth}{!}{%
\begin{tabular}{cccccccccccccc}
\toprule
\multirow{4}{*}{\textbf{Model}} & \multirow{4}{*}{\textbf{Language}}
  & \multicolumn{6}{c}{\textbf{English-Lang}}
  & \multicolumn{6}{c}{\textbf{Lang-English}} \\
\cmidrule(lr){3-8}\cmidrule(lr){9-14}
&  & \multicolumn{3}{c}{\textbf{nDCG@5}} & \multicolumn{3}{c}{\textbf{nDCG@10}}
   & \multicolumn{3}{c}{\textbf{nDCG@5}} & \multicolumn{3}{c}{\textbf{nDCG@10}} \\
\cmidrule(lr){3-5}\cmidrule(lr){6-8}\cmidrule(lr){9-11}\cmidrule(lr){12-14}
& 
  & \textbf{Base} & \textbf{InfoNCE} & \textbf{CLEAR}
  & \textbf{Base} & \textbf{InfoNCE} & \textbf{CLEAR}
  & \textbf{Base} & \textbf{InfoNCE} & \textbf{CLEAR}
  & \textbf{Base} & \textbf{InfoNCE} & \textbf{CLEAR} \\
\midrule

\multirow{5}{*}{bge-m3}
  & ar  & 91.84 & 92.29 & \textbf{93.19}
        & 92.24 & 92.53 & \textbf{93.46}
        & 91.91 & 94.12 & \textbf{94.26}
        & 92.38 & 94.52 & \textbf{94.62} \\
  & zh  & 93.85 & \textbf{94.22} & 94.21
        & 94.04 & 94.50 & \textbf{94.58}
        & 92.74 & 93.53 & \textbf{93.98}
        & 93.18 & 93.83 & \textbf{94.26} \\
  & es  & 95.98 & 95.12 & \textbf{96.04}
        & 96.14 & 95.37 & \textbf{96.29}
        & 95.80 & 95.68 & \textbf{96.03}
        & 95.96 & 95.95 & \textbf{96.26} \\
  & de  & 95.29 & 95.06 & \textbf{95.65}
        & 95.52 & 95.22 & \textbf{95.81}
        & \textbf{95.89} & 95.29 & 95.81
        & \textbf{96.17} & 95.46 & 95.91 \\
  & ru  & 95.29 & 95.01 & \textbf{95.40}
        & 95.58 & 95.15 & \textbf{95.73}
        & 94.17 & 94.26 & \textbf{94.50}
        & 94.57 & 94.56 & \textbf{94.88} \\
\midrule

\multirow{5}{*}{multilingual-e5}
  & ar  & 86.36 & 87.23 & \textbf{89.19}
        & 87.29 & 87.70 & \textbf{89.79}
        & 90.82 & 91.64 & \textbf{92.33}
        & 91.22 & 92.14 & \textbf{92.76} \\
  & zh  & 88.65 & 90.39 & \textbf{91.64}
        & 89.60 & 90.78 & \textbf{92.06}
        & 90.47 & 92.09 & \textbf{92.96}
        & 91.02 & 92.39 & \textbf{93.21} \\
  & es  & \textbf{96.01} & 92.79 & 93.80
        & \textbf{96.15} & 93.04 & 93.96
        & \textbf{96.25} & 93.73 & 94.64
        & \textbf{96.41} & 94.01 & 94.80 \\
  & de  & \textbf{94.84} & 91.97 & 93.50
        & \textbf{95.18} & 92.22 & 93.81
        & \textbf{95.41} & 93.50 & 94.26
        & \textbf{95.63} & 93.75 & 94.39 \\
  & ru  & \textbf{92.50} & 91.09 & 91.89
        & \textbf{93.06} & 91.51 & 92.28
        & 92.88 & 92.46 & \textbf{93.02}
        & 93.22 & 92.87 & \textbf{93.62} \\
\midrule

\multirow{5}{*}{gte-multilingual}
  & ar  & 86.65 & 87.48 & \textbf{88.76}
        & 87.44 & 88.20 & \textbf{89.35}
        & 91.29 & 92.61 & \textbf{93.23}
        & 91.89 & 93.06 & \textbf{93.63} \\
  & zh  & 93.70 & 94.07 & \textbf{94.67}
        & 93.98 & 94.28 & \textbf{94.86}
        & 91.61 & 92.78 & \textbf{93.47}
        & 92.07 & 93.11 & \textbf{93.83} \\
  & es  & 95.85 & 95.05 & \textbf{95.94}
        & 96.14 & 95.29 & \textbf{96.13}
        & 95.48 & 95.65 & \textbf{96.15}
        & 95.70 & 95.84 & \textbf{96.31} \\
  & de  & 94.81 & 93.70 & \textbf{94.82}
        & 95.11 & 94.06 & \textbf{95.16}
        & 94.47 & 94.93 & \textbf{95.84}
        & 94.75 & 95.24 & \textbf{96.01} \\
  & ru  & \textbf{94.23} & 93.29 & 94.18
        & 94.38 & 93.68 & \textbf{94.47}
        & 93.82 & 94.12 & \textbf{94.63}
        & 94.24 & 94.54 & \textbf{95.07} \\
\midrule

\multirow{5}{*}{jina-v3}
  & ar  & 90.01 & 92.72 & \textbf{93.66}
        & 90.58 & 93.12 & \textbf{93.90}
        & 92.83 & 94.76 & \textbf{94.84}
        & 93.21 & 95.02 & \textbf{95.05} \\
  & zh  & 92.32 & 94.46 & \textbf{94.67}
        & 92.65 & 94.65 & \textbf{94.87}
        & 92.37 & 94.84 & \textbf{95.21}
        & 92.79 & 95.06 & \textbf{95.34} \\
  & es  & 94.37 & 95.91 & \textbf{96.02}
        & 94.76 & 96.13 & \textbf{96.13}
        & 95.91 & 96.44 & \textbf{96.64}
        & 96.12 & 96.61 & \textbf{96.78} \\
  & de  & 94.52 & 95.87 & \textbf{95.98}
        & 94.69 & 96.05 & \textbf{96.14}
        & 95.15 & 96.26 & \textbf{96.54}
        & 95.30 & 96.34 & \textbf{96.62} \\
  & ru  & 93.77 & 95.81 & \textbf{95.81}
        & 93.97 & 95.98 & \textbf{95.98}
        & 94.06 & 95.50 & \textbf{95.59}
        & 94.38 & 95.77 & \textbf{95.81} \\
\bottomrule
\end{tabular}%
}
\caption{Results on all languages under the multilingual training setup in XQuAD.}
\label{tab:xquad_full_multi}
\end{table*}

\end{document}